\documentclass[runningheads]{llncs}

\usepackage[mobile]{eccv}

\usepackage{eccvabbrv}

\usepackage{graphicx}
\usepackage{booktabs}

\usepackage[accsupp]{axessibility}  % Improves PDF readability for those with disabilities.

\usepackage{hyperref}

\usepackage[utf8]{inputenc} % allow utf-8 input
\usepackage[T1]{fontenc}    % use 8-bit T1 fonts
\usepackage{hyperref}       % hyperlinks
\usepackage{url}            % simple URL typesetting
\usepackage{amsfonts}       % blackboard math symbols
\usepackage{nicefrac}       % compact symbols for 1/2, etc.
\usepackage{epsfig}
\usepackage{graphicx}
\usepackage{amsmath}
\usepackage{amssymb}
\def\eg{{\emph{e.g.}}}

\usepackage{graphicx}
\usepackage{amsmath}
\usepackage{amssymb}
\usepackage{pifont} % colors
\usepackage{amsmath}
\usepackage{amssymb}
\usepackage{graphicx}
\usepackage{array}
\usepackage{comment}
\usepackage{epsfig}
\usepackage{comment}
\usepackage{url}
\usepackage{verbatim}
\usepackage{multirow}
\usepackage{colortbl}
\usepackage{bbding}
\usepackage{pifont}
\usepackage{wasysym}
\usepackage{amssymb}
\usepackage{graphicx,animate}
\usepackage[normalem]{ulem}
\usepackage{tikz}
\newcommand*{\circled}[1]{\lower.7ex\hbox{\tikz\draw (0pt, 0pt)%
    circle (.4em) node {\makebox[1em][c]{\small #1}};}}

\def\eg{\emph{e.g.}}

\def\eg{{\emph{e.g.}}}

\def\Pi{{\mathbf{M}_{i}}}

\def\ee{{\mathbf{E}}}

\usepackage{makecell}

\usepackage[ruled,linesnumbered,boxed]{algorithm2e}
\usepackage{orcidlink}

\begin{document}

% \title{An Implicit Forward Solution to Electromagnetic Inverse Scattering Problems} 

\title{Imaging Interiors: An Implicit Solution to Electromagnetic Inverse Scattering Problems} 

\titlerunning{An Implicit Solution to Electromagnetic Inverse Scattering Problems}

\author{Ziyuan Luo\inst{1}\orcidlink{0000-0003-1580-9809} \and
Boxin Shi\inst{2, 3}\orcidlink{0000-0001-6749-0364} \and
Haoliang Li\inst{4}\orcidlink{0000-0002-8723-8112} \and
Renjie Wan\inst{1}\thanks{Corresponding author. This work was done at Renjie’s Research Group at the Department of Computer Science of Hong Kong Baptist University.}\orcidlink{0000-0002-0161-0367}}

\authorrunning{Z.~Luo et al.}

\institute{Department of Computer Science, Hong Kong Baptist University \and
National Key Laboratory for Multimedia Information Processing, School of Computer Science, Peking University \and
National Engineering Research Center of Visual Technology, School of Computer Science, Peking University \and
Department of Electrical Engineering, City University of Hong Kong \\
\email{ziyuanluo@life.hkbu.edu.hk}, \email{shiboxin@pku.edu.cn}, \email{haoliang.li@cityu.edu.hk}, \email{renjiewan@hkbu.edu.hk}}

\maketitle

\begin{abstract}
  Electromagnetic Inverse Scattering Problems (EISP) have gained wide applications in computational imaging. By solving EISP, the internal relative permittivity of the scatterer can be non-invasively determined based on the scattered electromagnetic fields. Despite previous efforts to address EISP, achieving better solutions to this problem has remained elusive, due to the challenges posed by inversion and discretization. This paper tackles those challenges in EISP via an implicit approach. By representing the scatterer's relative permittivity as a continuous implicit representation, our method is able to address the low-resolution problems arising from discretization. Further, optimizing this implicit representation within a forward framework allows us to conveniently circumvent the challenges posed by inverse estimation. Our approach outperforms existing methods on standard benchmark datasets. Project page: \href{https://luo-ziyuan.github.io/Imaging-Interiors/}{https://luo-ziyuan.github.io/Imaging-Interiors}.
  \keywords{Electromagnetic inverse scattering problems \and Implicit neural representations \and Computational photography}
\end{abstract}

\section{Introduction}
\label{sec:intro}
As electromagnetic waves can penetrate objects' surfaces, they play a key role in non-invasive imaging. Compared with modalities like X-ray and MRI, electromagnetic waves provide a potentially low-cost and safe approach~\cite{bevacqua2021millimeter, o2018microwave, geng2023dream} for non-invasive imaging.

\begin{figure}[!ht]
  \centering
  \includegraphics[width=\linewidth]{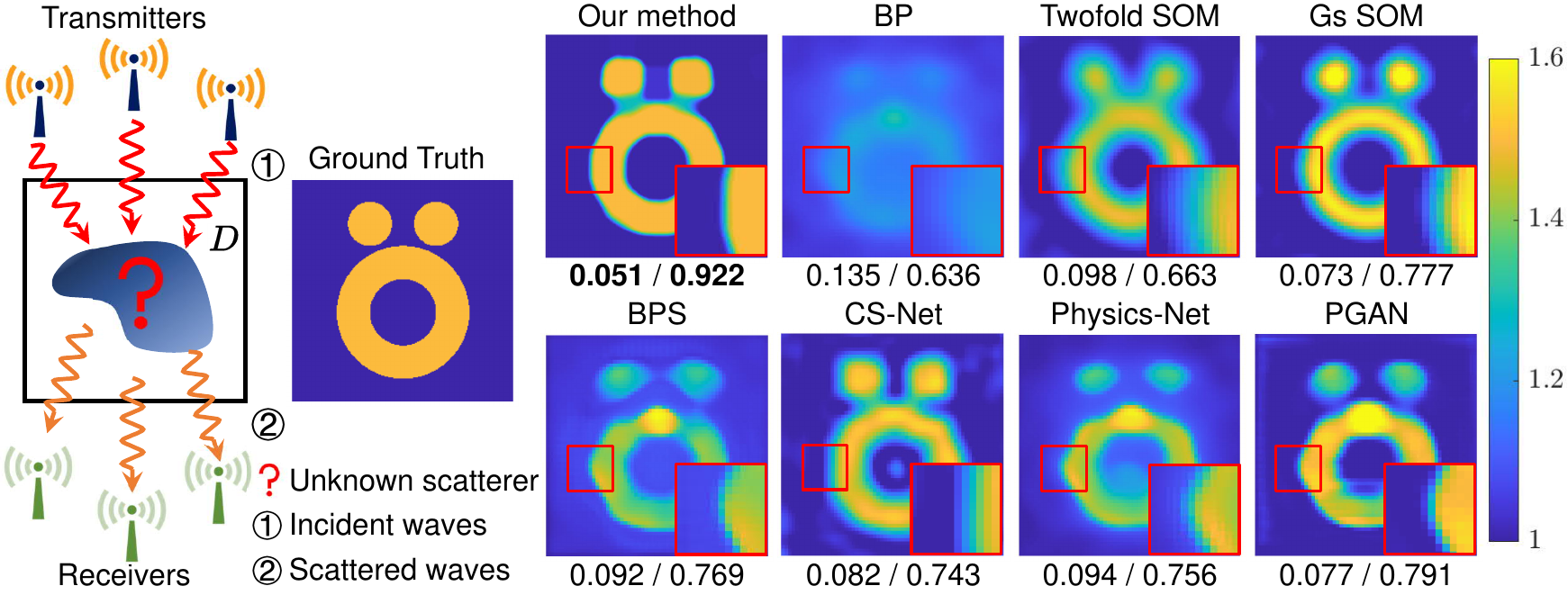}
  \caption{The results of a standard test case~\cite{belkebir1996using,wei2018deep} in EISP. In an EISP system, the scatterer in the enclosed space $D$ is first illuminated by incoming electromagnetic waves emitted by transmitters and generates scattered fields. Then, the scattered fields measured by receivers are used to determine the scatterer's relative permittivity. We show results obtained by our method, {BP}~\cite{belkebir2005superresolution}, Twofold SOM~\cite{zhong2009twofold}, Gs SOM~\cite{chen2009subspace}, BPS~\cite{wei2018deep}, CS-Net~\cite{sanghvi2019embedding}, Physics-Net~\cite{liu2022physics}, and {PGAN}~\cite{9468919}. The pixel values in the images indicate the values of the relative permittivity. RRMSE/SSIM values are shown below each figure.}
  \label{fig:fig1}
\end{figure}

Non-invasive imaging via electromagnetic waves needs to be completed by solving the Electromagnetic Inverse Scattering Problems (EISP). The typical scenario and the associated physical quantities of EISP are displayed in~\cref{fig:fig1}. In EISP, we need to infer the distribution of the scatterer's relative permittivity from the measured scattered fields~\cite{nikolova2011microwave}. Then, the values of relative permittivity can be visualized to image the objects' internal structures. 

However, solving EISP has never been an easy task. The first challenge comes from the \textbf{inverse} estimation process, where the scatterer's relative permittivity is derived from the measured scattered fields. The multiple scattering effects inherent to the EISP further significantly complicate this inverse estimation~\cite{chen2018computational, zhong2016new, li2018deepnis}. Furthermore, continuous spaces are usually discretized into finite elements or grids to facilitate numerical computations for electromagnetic computations. This process of \textbf{discretization} inherently leads to a loss of details and a decrease in image resolution for EISP~\cite{cui2004study, nordebo2007fisher}. This consequently poses challenges in accurately distinguishing internal small objects in close proximity.

The scattering mechanism is crucial in solving EISP~\cite{liu2022physics, chen2009subspace}. Recent deep-learning-based methods~\cite{li2018deepnis, wei2018deep} first obtain a rough image using traditional algorithms (\eg,~BP~\cite{belkebir2005superresolution} shown in~\cref{fig:fig1}) and then refine this rough result using image-to-image translation networks. However, with the imaging process being divided into two distinct stages, the measured physical data are overlooked in the second phase, inevitably resulting in images of inferior quality (\eg,~BPS~\cite{wei2018deep} and {PGAN}~\cite{9468919} shown in~\cref{fig:fig1}). Besides, as the relative permittivity distributions can vary greatly across different objects, a single network may be insufficient to reflect these differences. Therefore, a more appropriate approach is to optimize for each object while thoroughly considering the scattering mechanisms. Traditional methods like Twofold SOM~\cite{zhong2009twofold} and Gs SOM~\cite{chen2009subspace} achieve this goal by discretizing continuous relative permittivity values into discrete variables expressed as to-be-optimized matrices~\cite{chew1990reconstruction, zhong2009twofold, chen2009subspace}. However, they cannot address the low-resolution issues that arise from discretization, as shown in~\cref{fig:fig1}. Moreover, these methods have difficulties in handling sparse measurement data, which usually leads to low reconstruction quality. Thus, we need a more powerful representation scheme beyond the discrete matrix form.

Since the scatterer’s relative permittivity distribution used for imaging is intimately related to the spatial position~\cite{chew1990reconstruction}, the envisioned approach should indicate such location-dependent physical quantity. Moreover, for better imaging quality, the approach should have the capacity to eliminate the low-resolution curse made by discretization. Therefore, we propose to represent the scatterer's relative permittivity distribution via Implicit Neural Representations (INR), as INR has remarkable capability in modeling location-dependent relationships~\cite{nerf2020, sitzmann2020implicit}. Its flexibility in handling image resolutions~\cite{chen2022vinr, chen2021learning, hu2022learning} also helps to \textit{alleviate the curse made by discretization}. Besides, INR is known for its case-by-case optimization strategy, which can faithfully reflect each object's internal differences. In addition, INRs exhibit a strong capacity to recover information from incomplete data across a range of tasks~\cite{jagatap2019phase, nerf2020, 8579082}, which can address the difficulties caused by sparse measurement. We then optimize the INR of each object by making sure that this representation can produce results akin to the actual measurements. Such an optimization based on forward estimation can help to \textit{address the difficulties caused by inverse estimation}. Once the optimization settles down, we can access the relative permittivity values by using their corresponding locations and visualize them for imaging purposes.

Our solution is illustrated in~\cref{fig:framework}. We use the Multilayer Perceptron (MLP) as the backbone of INR. Specifically, two MLPs are used to represent the relative permittivity and the induced current. It allows us to bypass complex matrix inversions in optimization, largely reducing the computational cost. To optimize the implicit representations, we propose to use a data loss and a state loss by fully considering the relationships between various physical quantities in the scattering process. Meanwhile, a random sampling strategy is designed for comprehensive consideration of each spatial position during optimization. Our primary contributions can be summarized as follows:
\begin{itemize}
    \item We propose a solution to EISP based on forward estimation to avoid difficulties triggered by inverse estimation.

    \item We introduce the use of the learnable INR for the object's relative permittivity distribution to indicate the location-dependent quantity and achieve flexibility in imaging resolution.
    
    \item We explore a novel strategy that utilizes two separate MLPs for two physical quantities to avoid complex computations caused by matrix inversion.
\end{itemize}

The research for solving EISP has established standard benchmarks for evaluation~\cite{chen2018computational, wei2018deep, 9468919}. We strictly follow these standard benchmarks in our research, including system settings, datasets, and metrics.

\section{Related work} 
\noindent{\textbf{Electromagnetic inverse scattering problems (EISP).}}
Conventional approaches for EISP can be classified as non-iterative and iterative categories. Non-iterative methods tackle nonlinear issues by converting them into linear equations, such as the Born approximation~\cite{1132783}, Rytov approximation~\cite{devaney1981inverse, habashy1993beyond}, and back-propagation (BP) method~\cite{belkebir2005superresolution}. Since non-iterative methods are limited by reconstruction quality, other methods, such as the Subspace Optimization Method (SOM)~\cite{chen2009subspace, zhong2011fft, xu2017hybrid}, Contrast Source Inversion (CSI) Method~\cite{habashy1994simultaneous, van1999extended}, Distorted Born Iterative Method (DBIM)~\cite{gao2015sensitivity}, and other Newton-type methods~\cite{mojabi2010comparison, rubaek2007nonlinear}, are proposed to solve this problem in an iterative way. Despite their feasibility, their performance is highly sensitive to the initial guess, which undermines their robustness. Neural networks have recently been employed to construct nonlinear mappings linking the dispersed fields and the obscure constitutive attributes of scatterers~\cite{geng2021recent, li2024ifnet}. Most methods consider EISP in two stages~\cite{wei2018deep, li2018deepnis, zhang2020learning, xu2021fast}, where traditional approaches~\cite{1132783, belkebir2005superresolution} are first employed to obtain coarse images, and then the trained image-to-image neural networks are used for post-processing. However, due to the lack of consideration for scattering mechanisms, those methods encounter nearly all the difficulties faced by deep learning~\cite{li2024graph, pang2024heterogeneous}, which easily causes their post-processing efforts to fail. We consider an implicit forward approach to overcome the aforementioned difficulties by optimizing a trainable implicit representation within a forward process.

\noindent{\textbf{Implicit neural representations (INR).}}
Implicit neural representations usually employ an MLP to map from local coordinates to the associated values on that coordinates~\cite{sitzmann2020implicit, grattarola2022generalised, zhang2024ntinr}, such as intensity for images and videos, or occupancy value for 3D volumes. \textbf{The basics of INR are briefly explained below}. INR is a neural network $F_\theta: \mathbf{x} \mapsto F_\theta(\mathbf{x})$ that continuously maps the coordinates $\mathbf{x}$ to the corresponding quantity of interest. For the data expressed as a function $\Phi: \mathbf{x} \mapsto \Phi(\mathbf{x})$, the INR $F_\theta$ is a solution to $F_\theta(\mathbf{x}) - \Phi(\mathbf{x}) = 0$. Typically, the weight of INR, $\theta$, can be obtained through optimization. INR has shown its advantages in parameterizing geometry and learning priors over shapes, as demonstrated by DeepSDF~\cite{park2019deepsdf}, Occupancy Networks~\cite{mescheder2019occupancy}, and IM-Net~\cite{chen2019learning}.  A considerable amount of subsequent research has proposed volumetric rendering of 3D implicit representations, including Neural Radiance Fields (NeRF)~\cite{nerf2020} and its acceleration~\cite{muller2022instant, Chen2022ECCV}, quality enhancement~\cite{tang2024neural, cheng2024colorizing, zhu2023occlusion}, and copyright protection~\cite{luo2023copyrnerf}. In the realm of 2D image representation, CPPNs~\cite{stanley2007compositional} first proposed the use of neural networks to parameterize images implicitly. SIRENs~\cite{sitzmann2020implicit} proposed generalization across implicit representations of images via hypernetworks. X-Fields~\cite{bemana2020x} parameterizes the Jacobian of pixel position with respect to view, time, and illumination. In this work, we employ INR as the backbone to represent the scatterer's properties, which is able to characterize the complex spatial correlation in EISP and produce better imaging quality.

% \vspace{-10pt}
\section{Physical model of EISP}
\label{sec:smm}

As shown in~\cref{fig:fig1}, for an EISP system, the object to be imaged is defined as the unknown scatterer, and the enclosed space is represented by a square Region of Interest (ROI) denoted by $D$. Transmitters and receivers are positioned around $D$ to emit electromagnetic waves and measure the scattered electromagnetic fields, respectively. 

Generally, the data measurement process can be split into two stages. At the first stage, the induced current is excited as the electromagnetic waves from the transmitters interact with the scatterer, aka wave-scatterer interaction. Subsequently, the induced current serves as a radiation source, generating the scattered fields. The total electric fields \uline{within $D$} can be described by Lippmann-Schwinger equation~\cite{colton2013integral}, aka state equation, as follows:
\begin{equation}
E^{\mathrm{t}}(\mathbf{x})=E^{\mathrm{i}}(\mathbf{x})+k_0^2 \int_D g\left(\mathbf{x}, \mathbf{x}^{\prime}\right) J\left(\mathbf{x}^{\prime}\right) d \mathbf{x}^{\prime},\  \mathbf{x} \in D,\label{eq:state1}
\end{equation}
where $\mathbf{x}$ and $\mathbf{x}^{\prime}$ are the spatial coordinates. $E^{\mathrm{i}}$ is the incident electric fields directly from transmitters, while $E^{\mathrm{t}}$ is the total electric fields consisting of the original incident fields coming directly from
transmitters and the scattered fields coming from scatterers. $k_0$ is the wavenumber computed from the signal frequency, and $g$ is the free space Green's function. The relationship between the induced current $J$ and total electric fields  $E^{\mathrm{t}}$ can be expressed as follows:
\begin{equation}
J(\mathbf{x})=\xi(\mathbf{x}) E^t(\mathbf{x}). \label{eq:JE}
\end{equation}
In~\cref{eq:JE}, $\xi$ is the contrast defined as $    \xi(\mathbf{x})=\varepsilon_r(\mathbf{x})-1$, where $\varepsilon_r(\mathbf{x})$ denotes the relative permittivity of the unknown scatterer.

The second equation describes the scattered fields as a reradiation of the induced current~\cite{colton2013integral}, aka data equation, as follows:
\begin{equation}
E^{\text{s}}(\mathbf{x})=k_0^2 \int_D g\left(\mathbf{x}, \mathbf{x}^{\prime}\right) J\left(\mathbf{x}^{\prime}\right) d \mathbf{x}^{\prime}, \  \mathbf{x} \in S,\label{eq:data1}
\end{equation}
where $E^s$ is the scattered ﬁelds that can be measured by receivers \uline{at surface $S$} around the ROI $D$.

Since the digital signal analysis is only available on discrete variables\footnote{Discrete variables are denoted using \textbf{bold letters} in our paper}~\cite{mitra2001digital, taflove2005computational}, \cref{eq:state1} to \cref{eq:data1} are inevitably transformed to their discrete counterparts. Specifically, the ROI $D$ is discretized into $M \times M$ square subunits and the method of moments~\cite{peterson1998computational} is employed to obtain the discrete scattered fields. The relationship between discrete contrast $\boldsymbol{\xi}$ and discrete relative permittivity $\boldsymbol{\varepsilon}_r$ can be expressed as $\boldsymbol{\xi} = \boldsymbol{\varepsilon}_r - 1$. Then, \cref{eq:state1} can be reformulated as follows:
\begin{equation}
{\mathbf{E}}^{\mathrm{t}}={\mathbf{E}}^{\mathrm{i}}+{{\mathbf{G}}}_D \cdot {\mathbf{J}},\label{eq:d1}
\end{equation}
where ${\mathbf{E}}^{\mathrm{t}}$, ${\mathbf{E}}^{\mathrm{i}}$, and ${\mathbf{J}}$ are the discrete $E^{t}$, $E^{i}$, and $J$, respectively. ${{\mathbf{G}}}_D$ is discrete free space Green’s function in $D$. The discrete version of~\cref{eq:JE} can be represented as follows~\cite{9468919}:
\begin{equation}
{\mathbf{J}} = {\text{Diag}(\boldsymbol{\xi})} \cdot {\mathbf{E}}^{\mathrm{t}},
\label{eq:d3}
\end{equation}
where $\text{Diag}(\boldsymbol{\xi})$ is the diagonal matrix of contrast function.
Similarly, \cref{eq:data1} can be discretized as follows:
\begin{equation}
{\mathbf{E}}^{\mathrm{s}}={\mathbf{G}}_S \cdot {\mathbf{J}},
\label{eq:d2}
\end{equation}
where ${\mathbf{E}}^{\mathrm{s}}$ is the discrete $E^{s}$, and ${{\mathbf{G}}}_S$ is the discrete Green’s function to map contrast source ${\mathbf{J}}$ to scattered fields ${\mathbf{E}}^{\mathrm{s}}$.

The task of EISP is to infer the relative permittivity $\varepsilon_r$, a value closely related to $\boldsymbol{\xi}$, \textit{from the scattered fields $\ee^{\text{s}}$ measured by receivers}, and then visualize it in an image. In an EISP system, knowing only $\ee^{\text{i}}$ and $\ee^{\text{s}}$ makes it challenging to estimate $\boldsymbol{\xi}$ inversely from $\ee^{\text{s}}$ based on~\cref{eq:d2}. Furthermore, the inherent nonlinearity originating from ${{\mathbf{G}}}_D \cdot {\mathbf{J}}$ in~\cref{eq:d1} complicates the estimation process~\cite{cui2004study,chew1999waves}. Ultimately, $\varepsilon_r$, a variable with continuous characteristics, can only be derived using the aforementioned discrete equations, thereby compromising the imaging resolution. Consequently, an effective strategy is necessary to produce more accurate outcomes for EISP.

\section{Proposed implicit solution}

\begin{figure*}[!t]
  \centering
  \includegraphics[width=\linewidth]{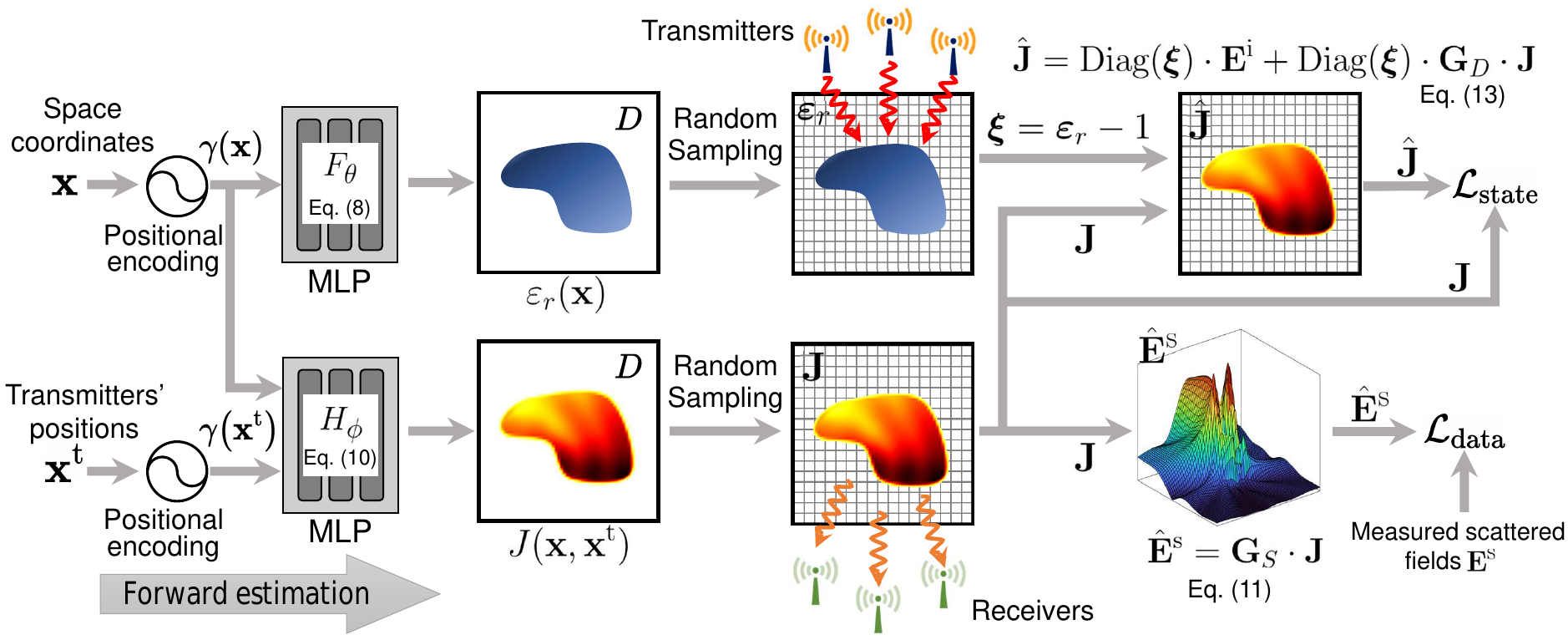}
  \caption{Overview of our implicit method. Two MLPs, $F_\theta$ and $H_\phi$, are used to implicitly represent relative permittivity $\varepsilon_r$ and induced current $J$, respectively. Random sampling is applied for comprehensive optimization. The predicted induced current $\hat{\mathbf{J}}$ is calculated by~\cref{eq:J} based on relative permittivity $\varepsilon_r$ queried from $F_\theta$ and induced current $J$ directly queried from $H_\phi$. Then the state loss $\mathcal{L}_{\text{state}}$ is calculated by comparing the predicted $\hat{\mathbf{J}}$ and directly queried $\mathbf{J}$. Besides, the directly queried induced current $\mathbf{J}$ is used to compute the scattered fields $\hat{\mathbf{E}}^{\text{s}}$ by~\cref{eq:E}. Data loss function $\mathcal{L}_{\text{data}}$ is constructed to evaluate the difference between predicted scattered fields $\hat{\mathbf{E}}^{\text{s}}$ and the measured values $\mathbf{E}^{\text{s}}$.}
  \label{fig:framework}
\end{figure*}
As depicted in \cref{fig:framework}, our fundamental idea is to solve the inverse scattering problem within a forward estimation process, which can be described by reformulating  \cref{eq:d1} to \cref{eq:d2} as follows:
\begin{equation}
{\mathbf{E}}^{\mathrm{s}} = {\mathbf{G}}_S \cdot \text{Diag}(\boldsymbol{\xi}) \cdot \left(\mathbf{I}-\mathbf{G}_D \text{Diag}(\boldsymbol{\xi})\right)^{-1} \cdot \mathbf{E}^{\text{i}}, \label{eq:inv}
\end{equation}
where $\mathbf{I}$ is the identity matrix. Contrary to previous approaches~\cite{1132783, belkebir2005superresolution} that treat $\boldsymbol{\xi}$ as unknown, we regard the contrast $\boldsymbol{\xi}$ as a known quantity, by establishing relative permittivity $\varepsilon_r$ involved in~\cref{eq:JE} as a learnable implicit representation. Then, we can calculate $\ee^{\text{s}}$ triggered by \cref{eq:inv}. During training, by 
contrasting the calculated $\ee^{\text{s}}$ with the measured true ones, the learnable representation can finally reach its optimum status. Since we do not aim to inversely express $\boldsymbol{\xi}$ as an explicit function of $\ee^{\text{s}}$, this optimization-in-forward approach can alleviate the challenges associated with inverse estimation and nonlinearity. The use of INR also ensures enhanced flexibility in imaging resolution. During inference, we can directly obtain the relative permittivity values by querying the representation with the corresponding coordinates. 

\subsection{Continuous representation for EISP}
\label{sec:ir}
% Scattered field can be regarded as the radiated field in free space by induced electromagnetic currents. (May be used somewhere)
\noindent{\textbf{Representation for relative permittivity.}} We use a multilayer perceptron (MLP) with parameters $\theta$ to map the continuous spatial coordinates $\mathbf{x}$ to its associating relative permittivity values, which can be briefly described as follows:
\begin{equation}
\varepsilon_r(\mathbf{x}) = F_\theta\left(\gamma(\mathbf{x})\right), \label{eq:epsilon}
\end{equation}
where $\varepsilon_r$\footnote{Due to its continuous property, we do not write it as bold font.} denotes the relative permittivity obtained by querying the spatial coordinates within the continuous representation $F_\theta$. We further apply the positional encoding for $\mathbf{x}$ to facilitate the fitting capability of the model as follows:
\begin{equation}
\gamma(\mathbf{x})=\left[\sin{\mathbf{x}}, \cos{\mathbf{x}}, \cdots, \sin{2^{\Omega-1}\mathbf{x}}, \cos{2^{\Omega-1}\mathbf{x}} \right]^\top,
\end{equation}
where $\Omega$ is a hyperparameter controlling the spectral bandwidth. 

To optimize this continuous representation, we apply a series of equations defined by the computation of forward estimation, and then compare the scattered fields incited by this representation with the measured true ones. As discussed in~\cref{sec:smm}, since the discretization is unavoidable during the forward computation, we only query discrete spatial coordinates during optimization to align with the discretization policy set by the method of moments~\cite{peterson1998computational}. Though we can directly perform the forward computation via \cref{eq:inv}, the computation complexity caused by its internal matrix inversion places a high demand on computing resources and potentially compromises the stability of problem-solving~\cite{golub2013matrix}.
    
\noindent{\textbf{Representation for induced current.}} To avoid the complexity caused by matrix inversion, instead of the explicit computation defined in~\cref{eq:inv}, we implicitly represent $\mathbf{J}$ as its continuous form using another MLP. By analyzing \cref{eq:inv}, the induced current is related to two variables: the discrete contrast $\boldsymbol{\xi}$ and the incident electric fields $\mathbf{E}^{\text{i}}$ caused by transmitters. To faithfully describe such correlation, we propose to consider the representation for $J$ as the mapping based on the spatial coordinates $\mathbf{x}$ and the transmitter's position $\mathbf{x}^{\text{t}}$, which can be defined as follows:
\begin{equation}
J(\mathbf{x}, \mathbf{x}^{\text{t}}) = H_\phi\left(\gamma(\mathbf{x}), \gamma(\mathbf{x}^{\text{t}})\right), \label{eq:J_inr}
\end{equation}
where $H_\phi$ is a continuous function consisting of fully connected networks.

\noindent{\textbf{Random spatial sampling.}}
Due to the necessity of discretization in numerical computations to perform forward process, we need to spatially query the representations $F_\theta$ and $H_\phi$ at some positions to obtain their discrete forms. A typical approach is to divide the ROI $D$ evenly into grids and deterministically sample the center location of each grid~\cite{sanghvi2019embedding, liu2022physics}. However, the representations would only be queried at a fixed discrete set of locations in this way~\cite{nerf2020}. To take a comprehensive consideration of each spatial position, we use a random sampling scheme where we partition ROI $D$ into $M \times M$ evenly-spaced grids, and the center of $(m,n)$ grid is $(x_m, y_n)$. Each sample location $(x_m^{\text{sample}}, y_n^{\text{sample}})$ is then randomly drawn from a Gaussian distribution: $x_m^{\text{sample}} \sim \mathcal N (x_m, \sigma^2), \quad y_n^{\text{sample}} \sim \mathcal N (y_n, \sigma^2)$, where $\sigma$ is a hyperparameter to control the dispersion level of sample points around their means. By probabilistically sampling each spatial position, this scheme can alleviate the overlook of specific locations.

\subsection{Forward calculation based optimization}
The continuous representations obtained before can estimate corresponding physical quantities using spatial coordinates, but optimizing them directly with estimated values is hard because obtaining their true values is difficult~\cite{wei2020uncertainty}. Therefore, we indirectly refine the representations denoted by $F_{\theta}$ and $H_\phi$ by proposing a data loss and a state loss while fully considering the physical relationships.

\noindent{\textbf{Data loss.}} Based on~\cref{sec:smm}, the wave-scatterer interaction leads to the induced current when the electromagnetic waves from transmitters interact with the scatterer. Then, the induced current generates the scattered fields, which can be measured by receivers. Thus, a straightforward way to refine the representations is to minimize the distance between the scattered fields computed from the representations and the true ones. With the discrete induced current $\mathbf{J}$ sampling from the continuous representation $H_{\phi}$ defined in~\cref{eq:J_inr}, we can obtain the predicted scattered fields  by reformulating the data equation in~\cref{eq:d2} as follows:
\begin{equation}
\hat{\mathbf{E}}^{\text{s}}_p={{\mathbf{G}}}_S \cdot {\mathbf{J}_p}, 
\label{eq:E}
\end{equation}
where ${{\mathbf{G}}}_S$ denotes the discrete Green's function, and $\mathbf{J}_p$ and $\hat{\mathbf{E}}^{\mathrm{s}}_p$ denotes the discrete induced current and predicted scattered fields inspired by the $p$-th transmitter, respectively.

We then define our data loss used to contrast the predicted scattered fields with the measured true ones as follows:
\begin{equation}
\mathcal{L}_{\text{data}} = {\sum_{p=1}^{N_{\text{t}}}\|\hat{\mathbf{E}}^{\text{s}}_p-{\mathbf{E}}^{\text{s}}_p\|^2},\label{eq:loss_data}
\end{equation}
where $\mathbf{E}^{\text{s}}_p$ is the true scattered fields measured by receivers, and $N_t$ denotes the number of all transmitters. The data loss enhances the optimization of the representation by accounting for discrepancies associated with all transmitters.

\noindent{\textbf{State loss.}} Although the data loss is straightforward, it cannot be used to optimize $F_{\theta}$. Since the induced current $\mathbf{J}_p$ is directly obtained by querying its representation $H_{\phi}$, \cref{eq:loss_data} does not contain any variables related to $F_{\theta}$. To effectively optimize $F_{\theta}$, we consider minimizing the mismatch emerging from the state equation defined in \cref{eq:d1} and \cref{eq:d3}. Specifically, by reformulating \cref{eq:inv}, an expression related to $p$-th transmitter can obtained as follows:
\begin{equation}
\hat{\mathbf{J}}_p = \text{Diag}(\boldsymbol{\xi})\cdot {\mathbf{E}}_p^{\text{i}}+\text{Diag}(\boldsymbol{\xi}) \cdot {{\mathbf{G}}}_D \cdot {\mathbf{J}}_p, \label{eq:J}
\end{equation}
where $\mathbf{J}_p$ denotes the discrete induced current sampling from $H_\phi$, $\boldsymbol{\xi}$ denotes the contrast, and $\hat{\mathbf{J}}_p$ indicates the induced current computed via the above mathematical correlation. 

Although $\mathbf{J}_p$ and $\hat{\mathbf{J}}_p$ originate from distinct sources, both represent the induced current values within the same spatial domain. If the same spatial coordinates are given, they should yield identical values irrespective of their generation sources. Therefore, we introduce a state loss to minimize the mismatch between $\mathbf{J}_p$ and $\hat{\mathbf{J}}_p$ as follows:
\begin{equation}
\mathcal{L}_{\text{state}} = {\sum_{p=1}^{N_{\text{t}}}\|\hat{\mathbf{J}}_p-{\mathbf{J}}_p\|^2},\label{eq:loss_state}
\end{equation}
where $N_{\text{t}}$ is the number of all transmitters. Since~\cref{eq:JE} has clearly defined the \uline{close correlation} between $\boldsymbol{\xi}$ and the permittivity values $\varepsilon_r$, the minimization of the state loss can ultimately optimize $F_{\theta}$ and $H_\phi$ used for representing the relative permittivity and induced current, respectively.

\noindent{\textbf{Overall loss.}}
The overall loss to train the relative permittivity representation $F_\theta$ and induced current representation $H_\phi$ can be obtained as follows:
\begin{equation}
\mathcal{L} = \lambda_{\text{data}}\mathcal{L}_{\text{data}} + \lambda_{\text{state}}\mathcal{L}_{\text{state}} + \lambda_{\text{TV}}\mathcal{L}_{\text{TV}}, 
\label{eq:loss_overall}
\end{equation}
where $\mathcal{L}_{\text{TV}}$ is a total variation loss for $\boldsymbol{\xi}$, and $\lambda_{\text{data}}$, $\lambda_{\text{state}}$, $\lambda_{\text{TV}}$ are hyperparameters to balance the loss functions.

\subsection{Implementation details}
We implement our method using PyTorch. Two eight-layer MLPs with $256$ channels and ReLU activations are used to predict the relative permittivity $\varepsilon_r$ and induced current $J$, respectively. Similar to the settings in previous methods based on INRs~\cite{nerf2020, sitzmann2020implicit}, positional encoding is applied to input positions before they are passed into the MLPs. The ROI $D$ is discretized into $64 \times 64$ while training. The hyperparameters for the overall loss are set as $\lambda_{\text{data}}=1.00$, $\lambda_{\text{state}}=1.00$, and $\lambda_{\text{TV}}=0.01$. We use the Adam optimizer with default values $\beta_1 = 0.9$, $\beta_2 = 0.999$, $\epsilon = 10^{-8}$, and a learning rate $5\times10^{-4}$ that decays following the exponential scheduler during the optimization. We optimize the model for $4$K iterations on a single NVIDIA V100 GPU for all the datasets.

\section{Experiments}
\subsection{Experiment setup}
\noindent{\textbf{Dataset.}} We strictly follow the established rules in EISP~\cite{wei2018deep, 9468919, liu2022physics} to set the dataset. We train and test our model on  \textbf{standard benchmarks} used for EISP.  1) Synthetic \textbf{Circular-cylinder dataset} is synthetically generated comprising $1200$ images of cylinders with random relative radius, number, and location and permittivity between $1$ and $1.5$. 2) Synthetic \textbf{MNIST dataset} contains grayscale images of handwritten digits. Similar to previous settings~\cite{wei2018deep, zhou2022deep}, we randomly select $1200$ of them to synthesize scatterers with relative permittivity values between $2$ and $2.5$ according to their corresponding pixel values. Following previous work to generate the above two synthetic datasets, we use $16$ transmitters and $32$ receivers equally placed on a circle for regular settings, and the data are generated numerically using the method of moments~\cite{peterson1998computational} with a $224 \times 224$ grid mesh to avoid inverse crime~\cite{colton1998inverse}. For sparse measurement experiments, we decrease the number of receivers from $32$ to $8$ to utilize only $25\%$ of the regular measurement data. 3) Real-world \textbf{Institut Fresnel's database} contains three different dielectric scenarios, namely FoamDielExt, FoamDielInt, and FoamTwinDiel. There are $8$ transmitters for FoamDielExt and FoamDielInt, $18$ transmitters for FoamTwinDiel, and 241 receivers for all the cases. As the wavelength of the emitted electromagnetic wave should be comparable to or smaller than the size of the target object, we set operating frequency $f = 400$ MHz on synthetic datasets, and $f = 5$ GHz on real-world dataset. 

\noindent{\textbf{Baselines.}} We maintain the same settings as in previous studies~\cite{wei2018deep,9468919,sanghvi2019embedding} to ensure a fair comparison. We compare our method with three traditional methods and four deep learning-based approaches: 1) \textbf{BP}~\cite{belkebir2005superresolution}: A traditional non-iterative inversion algorithm. 2) \textbf{Twofold SOM}~\cite{zhong2009twofold}: A traditional iterative minimization scheme by using SVD decomposition. 3) \textbf{Gs SOM}~\cite{chen2009subspace}: A traditional subspace-based optimization method by decomposing the operator of Green's function. 4) \textbf{BPS}~\cite{wei2018deep}: A CNN-based image translation method with an initial guess from the BP algorithm. 5) \textbf{CS-Net}~\cite{sanghvi2019embedding}: A CNN-based contrast source reconstruction scheme via subspace optimization.  6) \textbf{Physics-Net}~\cite{liu2022physics}: A CNN-based approach that incorporates physical phenomena during training. 7) \textbf{PGAN}~\cite{9468919}:  A CNN-based
approach using a generative adversarial network.

\noindent{\textbf{Evaluation methodology.}} We evaluate the quantitative performance of our method using PSNR, SSIM, and Relative Root-Mean-Square Error (RRMSE)~\cite{9468919}.
For PSNR and SSIM, a higher value indicates better performance. For RRMSE~\cite{9468919}, a lower value indicates better performance. RRMSE is a metric widely used in EISP defined as follows:
\begin{equation}
\mathrm{RRMSE}=({\frac{1}{M \times M} \sum_{m=1}^M \sum_{n=1}^M|\frac{\hat{\boldsymbol{\varepsilon}}_r(m, n)-\boldsymbol{\varepsilon}_r(m, n)}{\boldsymbol{\varepsilon}_r(m, n)}|^2})^{\frac{1}{2}},
\end{equation}
where $\boldsymbol{\varepsilon}_r(m, n)$  and $\hat{\boldsymbol{\varepsilon}}_r(m, n)$ are the true and reconstructed discrete relative permittivity of the unknown scatterers at location $(m, n)$, respectively, and $M \times M$ is the total number of subunits over the ROI $D$.

% \vspace{-10pt}
\begin{figure*}[thb]
  \centering
  \includegraphics[width=\linewidth]{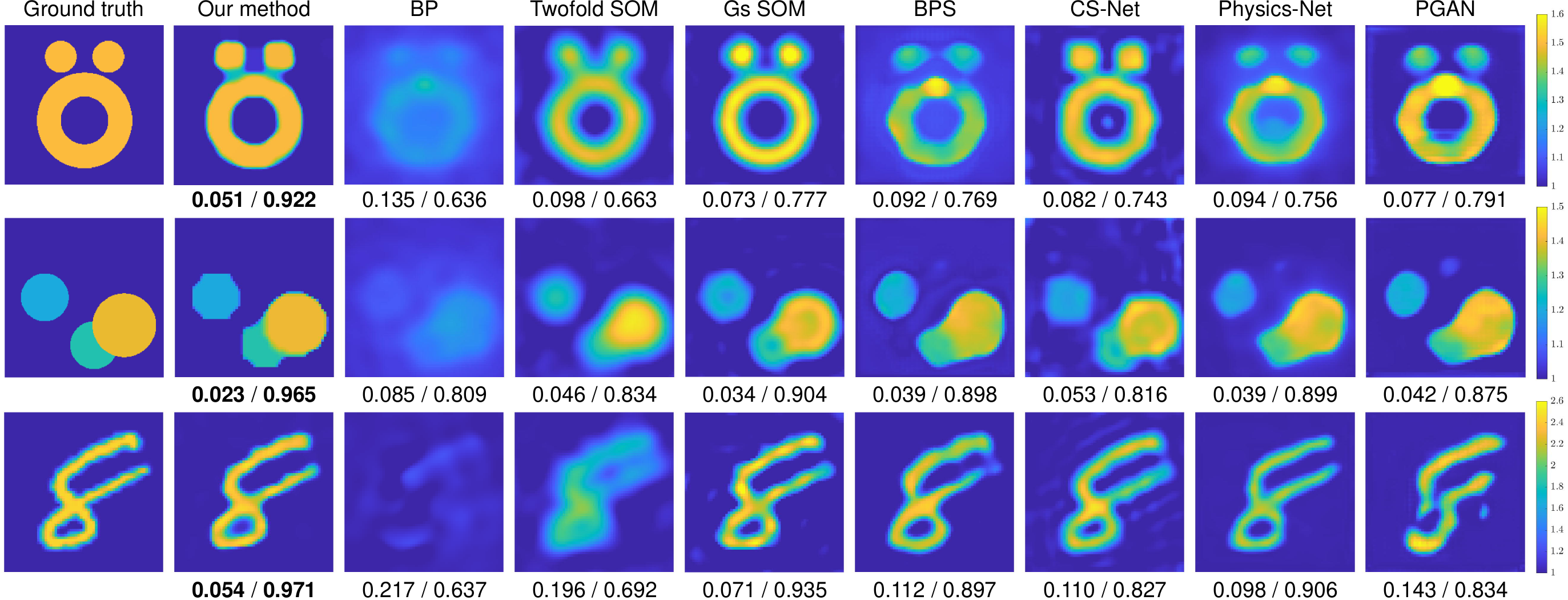}
  \caption{Samples obtained from synthetic Circular-cylinder dataset and MNIST dataset. From left to right: ground truth, results obtained using our method, {BP}~\cite{belkebir2005superresolution}, {Twofold SOM}~\cite{zhong2009twofold},  {Gs SOM}~\cite{chen2009subspace}, {BPS}~\cite{wei2018deep}, {CS-Net}~\cite{sanghvi2019embedding}, 
 {Physics-Net}~\cite{liu2022physics}, and {PGAN}~\cite{9468919}. The pixel values in the images indicate the values of the relative permittivity. RRMSE/SSIM values are shown below each figure. The first row is a standard test case~\cite{belkebir1996using, wei2018deep}, a well-known pattern for the evaluation of EISP methods.}
  \label{fig:results}
\end{figure*}

\begin{figure*}[thb]
  \centering
  \includegraphics[width=\linewidth]{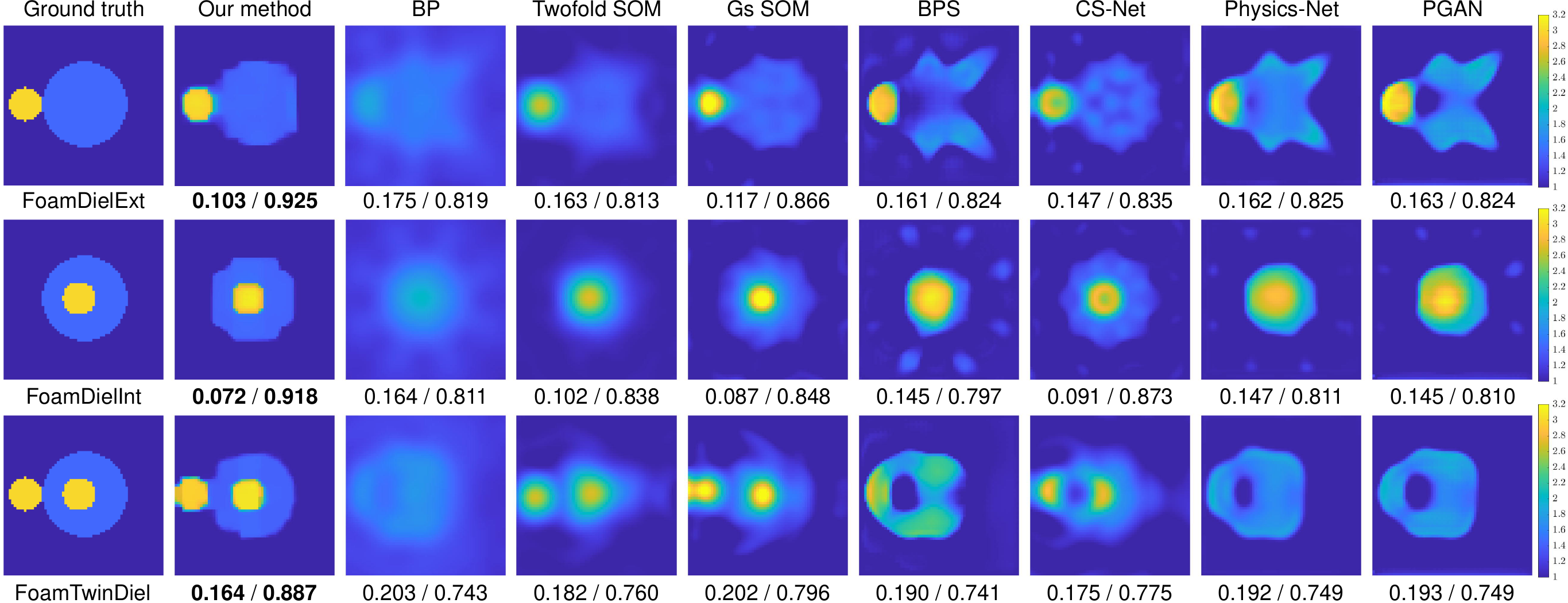 }
  \caption{Samples obtained from real-world Institut Fresnel’s database. From left to right: ground truth, results obtained using our method, {BP}~\cite{belkebir2005superresolution}, {Twofold SOM}~\cite{zhong2009twofold},  {Gs SOM}~\cite{chen2009subspace}, {BPS}~\cite{wei2018deep}, {CS-Net}~\cite{sanghvi2019embedding}, 
 {Physics-Net}~\cite{liu2022physics}, and {PGAN}~\cite{9468919}. The pixel values in the images indicate the values of the relative permittivity. RRMSE/SSIM values are shown below each figure.}
  \label{fig:real-results}
\end{figure*}

\subsection{Experiment results}
\noindent{\textbf{Qualitative results on synthetic data.}} We first compare the reconstruction quality visually using synthetic data including circular-cylinder dataset~\cite{wei2018deep} and MNIST dataset~\cite{deng2012mnist} against all baselines and the results are shown in~\cref{fig:results}. Our method achieves superior visual performance compared to all other baselines. Traditional methods, such as BP~\cite{belkebir2005superresolution} and Twofold SOM~\cite{zhong2009twofold}, can only recover the rough shape of the target and produce inaccurate results at some junction parts. Gs SOM~\cite{chen2009subspace} can reconstruct the relative permittivity of the targets with relatively high accuracy, but still has obvious visual defects. Though deep learning-based methods, such as BPS~\cite{wei2018deep}, CS-Net~\cite{sanghvi2019embedding}, Physics-Net~\cite{liu2022physics}, and PGAN~\cite{9468919}, can produce more accurate estimation, their visual qualities are still far below our proposed method. 

\noindent{\textbf{Qualitative results on real-world data.}}  We further test our algorithm on real-world data.  The results for all methods are shown in~\cref{fig:real-results}. The visual performance of our method remains superior when applied to real-world data. Conventional methods exhibit a diminished performance, which, when used as input, further negatively impacts the efficacy of deep learning-based approaches.

\begin{figure*}[!tb]
  \centering
  \includegraphics[width=\linewidth]{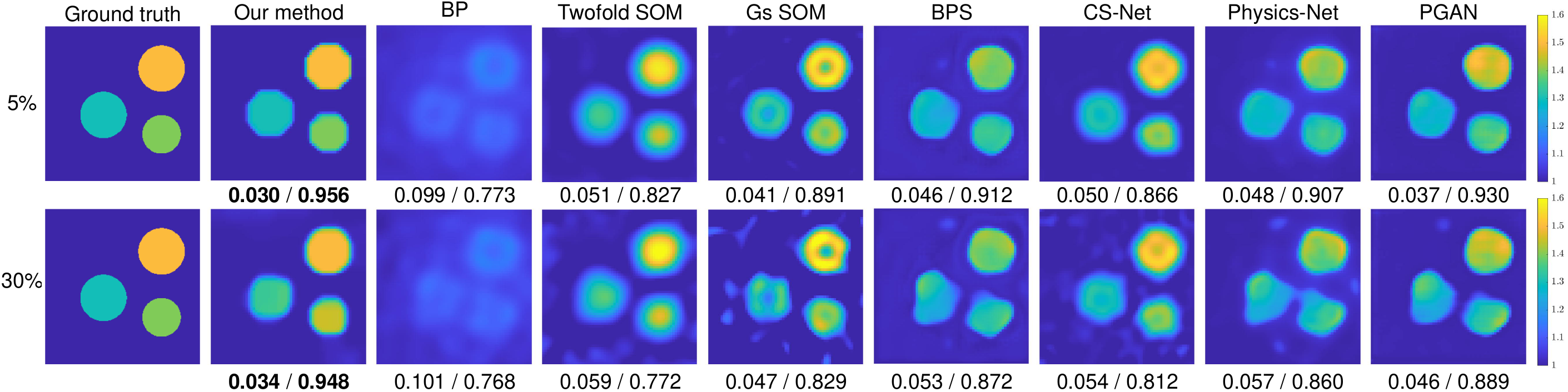}
  \caption{Samples obtained under $5\%$ and $30\%$ noise levels. From left to right: ground truth, results obtained using our method, {BP}~\cite{belkebir2005superresolution}, {Twofold SOM}~\cite{zhong2009twofold},  {Gs SOM}~\cite{chen2009subspace}, {BPS}~\cite{wei2018deep}, {CS-Net}~\cite{sanghvi2019embedding}, 
 {Physics-Net}~\cite{liu2022physics}, and {PGAN}~\cite{9468919}. The pixel values in the images indicate the values of the relative permittivity. RRMSE/SSIM values are shown below each figure.}
  \label{fig:noise}
\end{figure*}

% \vspace{-40pt}
\noindent{\textbf{Noise robustness.}} We also compare the robustness of the models. By adding different levels of noise to the received scattered fields signal, we use the noisy scattered fields signal to recover the scattered object. The results are shown in~\cref{fig:noise}. It can be observed that when large noise is added, the results of other baselines are greatly affected. However, our method can accurately reconstruct the scatterer under larger noise, demonstrating the superior robustness of the proposed method.

\begin{figure*}
  \centering
  \includegraphics[width=\linewidth]{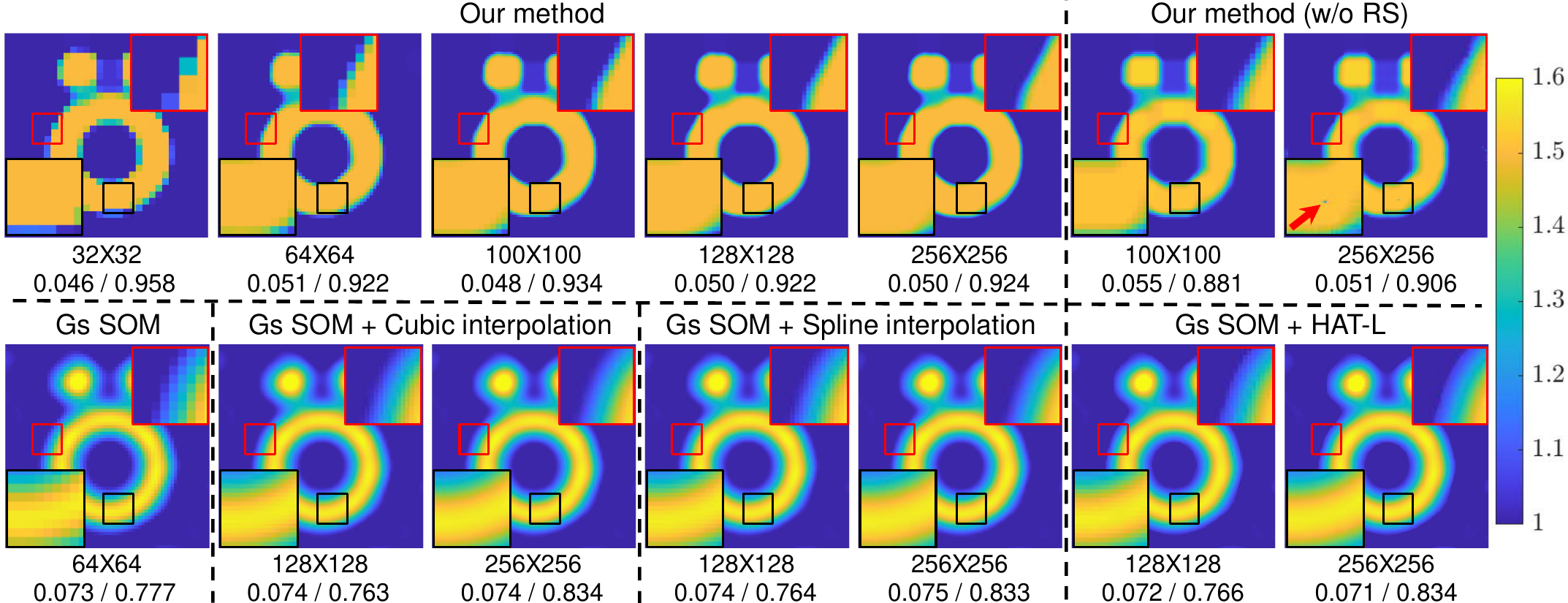}
  \caption{Comparison of resolution flexibility among different methods. Our method is evaluated against our method without random sampling (RS) scheme, {Gs SOM}~\cite{chen2009subspace}, Gs SOM + Cubic/Spline interpolation, and Gs SOM + HAT-L~\cite{chen2023activating}, under different resolution settings. All methods are implemented at an original resolution of $64 \times 64$. The pixel values in the images indicate the values of the relative permittivity. Resolution and RRMSE/SSIM values are shown below each figure.}
  \label{fig:multi-results}
\end{figure*}

% \vspace{-10pt}
\noindent{\textbf{Evaluation for flexible resolution.}} The continuity of implicit function allows INR to interpolate and infer at arbitrary points in the input space. This characteristic ensures flexible resolution during the imaging process. As depicted in~\cref{fig:multi-results}, we employ a fixed resolution of $64\times 64$ during training and acquire images of varying resolutions by sampling the INR at different scales. The use of the random sampling scheme can avoid artifacts in the results. Compared to post-super-resolution methods, like Gs SOM + HAT-L~\cite{chen2023activating}, our method can achieve higher quality results. The resolution flexibility of INR paves the way for more in-depth analysis of reconstructed images at various resolution levels.

\noindent{\textbf{Quantitative results.}} The quantitative results of the reconstruction quality in~\cref{tab:result} further validate the proposed method. Higher PSNR and SSIM values suggest our method accurately recovers object shapes and effectively maintains detailed structural information. Lower relative error values indicate our method's predictions of relative permittivity distribution are more accurate compared to other methods.

\begin{table}[!thb]
\caption{Quantitative evaluation compared with {BP}~\cite{belkebir2005superresolution}, {Twofold SOM}~\cite{zhong2009twofold}, {Gs SOM}~\cite{chen2009subspace}, {BPS}~\cite{wei2018deep}, {CS-Net}~\cite{sanghvi2019embedding}, 
 {Physics-Net}~\cite{liu2022physics}, and {PGAN}~\cite{9468919}. Best results are shown in \textbf{bold}.}
 \label{tab:result}
 \centering
 \resizebox{\textwidth}{!}{
\begin{tabular}{l|ccc|ccc|ccc|ccc|ccc}
\hline
\rule{0pt}{10pt}
            & \multicolumn{3}{c|}{Circular dataset (5\%)} & \multicolumn{3}{c|}{Circular dataset (30\%)} & \multicolumn{3}{c|}{MNIST dataset (5\%)} & \multicolumn{3}{c|}{MNIST dataset (30\%)} & \multicolumn{3}{c}{Institut Fresnel's database} \\ \cline{2-16} \rule{0pt}{10pt}
            %\cmidrule{2-16} 
            & RRMSE         & SSIM         & PSNR         & RRMSE         & SSIM          & PSNR         & RRMSE        & SSIM        & PSNR        & RRMSE        & SSIM         & PSNR        & RRMSE          & SSIM           & PSNR           \\ \hline \rule{-2pt}{10pt}
Proposed    & \textbf{0.016}         & \textbf{0.968}        & \textbf{39.20}        & \textbf{0.027}         & \textbf{0.938}         & \textbf{34.15}        & \textbf{0.017}        & \textbf{0.972}       & \textbf{37.90}       & \textbf{0.044}        & \textbf{0.893}        & \textbf{29.70}       & \textbf{0.127}          & \textbf{0.897}          & \textbf{27.06}          \\
BP          & 0.048         & 0.916        & 28.96        & 0.049         & 0.914         & 28.94        & 0.171        & 0.750       & 20.02       & 0.174        & 0.744        & 19.87       & 0.180          & 0.792          & 17.74          \\
Twofold SOM & 0.035         & 0.917        & 32.27        & 0.040         & 0.880         & 31.13        & 0.137        & 0.770       & 22.35       & 0.142        & 0.745        & 22.13       & 0.149          & 0.804          & 23.87          \\
Gs SOM      & 0.034         & 0.926        & 32.74        & 0.035         & 0.916         & 32.43        & 0.101        & 0.853       & 24.63       & 0.113        & 0.811        & 23.73       & 0.135          & 0.837          & 24.95          \\
BPS         & 0.027         & 0.964        & 34.20        & 0.034         & 0.928         & 32.17        & 0.098        & 0.912       & 25.68       & 0.133        & 0.859        & 23.21       & 0.166          & 0.787          & 18.86          \\
CS-Net      & 0.024              & 0.900          & 33.75             & 0.035              & 0.810              & 31.27             &  0.179            &  0.788           &  25.37           &    0.227          &          0.724    &   24.15          & 0.137          & 0.828          & 24.87          \\
Physics-Net & 0.024         & 0.945        & 37.20        & 0.031         & 0.921         & 33.53        & 0.079        & 0.938       & 27.74       & 0.113        & 0.889        & 24.44       & 0.168          & 0.795          & 18.42          \\
PGAN        & 0.021         & 0.957        & 36.89        & 0.030         & 0.933         & 33.03        & 0.090        & 0.918       & 26.09       & 0.124        & 0.873        & 23.64       & 0.167          & 0.794          & 18.46          \\ \hline
\end{tabular}
}
% \vspace{-5pt}
\end{table}

\begin{figure}[!tb]
  \centering
  \includegraphics[width=\linewidth]{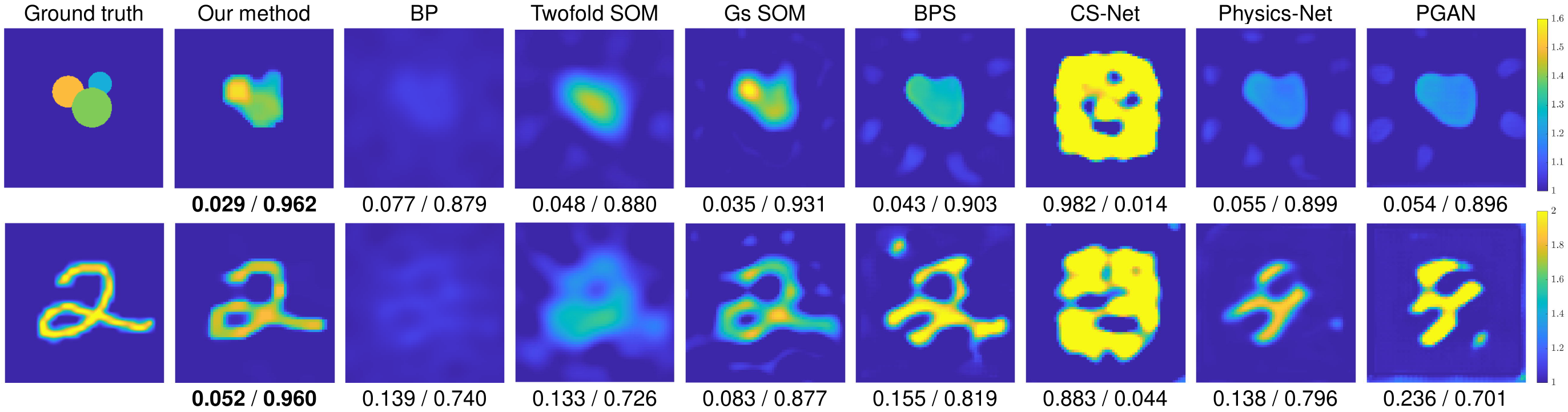}
  \caption{The results for sparse measurement. From left to right: ground truth, results obtained using our method, {BP}~\cite{belkebir2005superresolution}, {Twofold SOM}~\cite{zhong2009twofold},  {Gs SOM}~\cite{chen2009subspace}, {BPS}~\cite{wei2018deep}, {CS-Net}~\cite{sanghvi2019embedding}, 
 {Physics-Net}~\cite{liu2022physics}, and {PGAN}~\cite{9468919}. The pixel values in the images indicate the values of the relative permittivity. RRMSE/SSIM values are shown below each figure.}
  \label{fig:sparse}
\end{figure}

% \noindent{\textbf{Ablation study.}}
% \vspace{-10pt}
\noindent{\textbf{Results for sparse measurement.}} We use $25\%$ of the standard measurement data to conduct an experiment~\cite{wei2018deep, sanghvi2019embedding}, and the results are shown in~\cref{fig:sparse}. Our method, leveraging INR, shows superior capability in reconstructing the object's interior from sparse measurements compared with other methods. This also demonstrates that INR is more suitable for representing the relative permittivity distribution of the object compared to other matrix-based representations like {Twofold SOM}~\cite{zhong2009twofold} and {CS-Net}~\cite{sanghvi2019embedding}.

% \vspace{-60pt}

\begin{figure}[!tb]
  \centering
  \includegraphics[width=\linewidth]{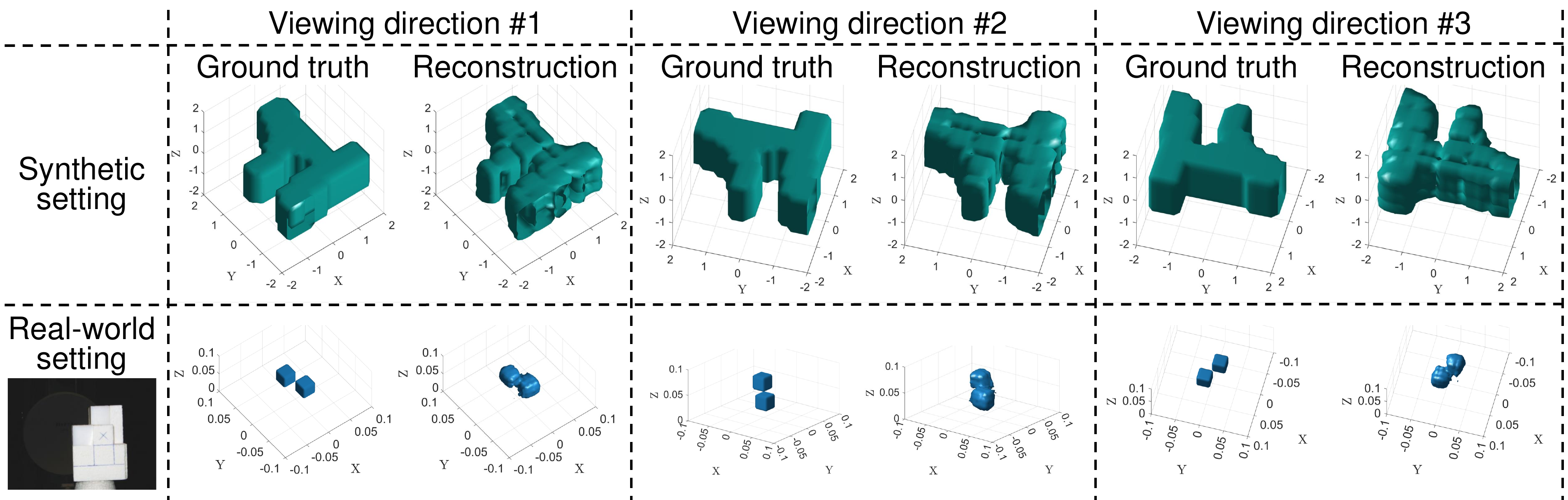}
  \caption{The results for 3D scenarios on 3D MINIST dataset~\cite{3Dmnist} and real-world dataset~\cite{geffrin2009continuing}. We show two example cases with three viewing directions. We use isosurface to visualize the ground truths and reconstructed results of the spatial distribution of relative permittivity.}
  \label{fig:3D}
\end{figure}

\noindent{\textbf{Results for 3D scenarios.}} Our method can naturally generalize to 3D scenarios. In 3D scenarios, we utilize 3D Green's functions in~\cref{eq:d1} and~\cref{eq:d2}, and perform integration over a 3D region of interest. We augment the input dimension of representations for relative permittivity and induced current to align with the 3D coordinates, while keeping the remaining structures unchanged. We collect the 3D scattering data from the synthetic 3D MNIST dataset~\cite{3Dmnist} and real-world dataset~\cite{geffrin2009continuing}. In the synthetic experiments, we employ $40$ transmitters and $160$ receivers arranged around a unit cube. The results shown in~\cref{fig:3D} demonstrate that our method can reconstruct 3D objects by solving EISP.

\noindent{\textbf{Time analysis.}} Our method, BP~\cite{belkebir2005superresolution}, Twofold SOM~\cite{zhong2009twofold}, and Gs SOM~\cite{chen2009subspace} are case-specific methods that do not require additional training time. The BPS~\cite{wei2018deep}, CS-Net~\cite{sanghvi2019embedding}, Physics-Net~\cite{liu2022physics}, and PGAN~\cite{9468919} are deep-learning methods that necessitate model training. We count the time for optimization procedure in inference time, and our method outperforms traditional case-specific methods and CS-Net~\cite{sanghvi2019embedding}. Although the inference of other deep-learning methods is very fast, their training processes are time-consuming. Additionally, we may be able to reduce the optimization time through modifying the INR structure~\cite{muller2022instant} or improving the training process~\cite{zhang2023nonparametric, zhang2024nonparametric}.

% \vspace{-10pt}

\subsection{Ablation study}
Our approach mainly consists of three parts: the representation for relative permittivity, representation for induced current, and random sampling. Since both the data loss $\mathcal{L}_{\text{data}}$ and the state loss $\mathcal{L}_{\text{state}}$ are necessary for our method, we cannot easily remove any of them. We first use a single representation for relative permittivity to replace the two MLPs in our original approach. Both models are trained on an NVIDIA V100 GPU. From the results shown in~\cref{tab:ablation}, the model with a single MLP takes a longer iteration time to achieve similar performance. We further remove the random sampling scheme and adopt a uniform sampling approach. As shown in the last two columns in~\cref{fig:multi-results}, results without random sampling have lower metric values, and the high-resolution results occasionally meet obvious artifacts.

\begin{table}[!htb]
%  \centering
%  \makeatletter\def\@captype{table}
 \caption{Ablation results compared with single representation for relative permittivity. }
 \centering
 \resizebox{.85\linewidth}{!}{
 \begin{tabular}{l|ccccc}
    \toprule 
         & Parameter counts{$\downarrow$} & Time each iteration{$\downarrow$} & RRMSE{$\downarrow$} & SSIM{$\uparrow$} & PSNR{$\uparrow$} \\
         \midrule
    %\midrule
    Proposed & 1,019,139 &\textbf{117 ms} & \textbf{0.038} & \textbf{0.909} & \textbf{30.36}\\ \midrule
    Single representation & \textbf{493,313} & 289 ms & 0.053 & 0.876 & 27.32 \\
    \bottomrule
    %\bottomrule
\end{tabular}
}
\label{tab:ablation}
\end{table}

\section{Conclusions}
An implicit solution is presented to the EISP in this paper. We propose implicitly representing the unknown scatterer's relative permittivity distribution as a trainable representation and optimizing these representations within a forward estimation process. A two-MLP strategy is employed to reduce the computation complexity. The experimental results on synthetic and real-world data demonstrate the promising performance of our proposed method. We will further collaborate with other parties to consider our solution in different scenarios, such as medical diagnostics, in more challenging conditions.

\section*{Acknowledgments}
This research is supported by the National Natural Science Foundation of China under Grant No. 62302415, Guangdong Basic and Applied Basic Research Foundation under Grant No. 2022A1515110692, 2024A1515012822, and the Blue Sky Research Fund of HKBU under Grant No. BSRF/21-22/16. It is also supported by National Natural Science Foundation of China under Grant No. 62136001, 62088102. It is also supported by Changsha Technology Fund under its grant No. KH2304007.

\bibliographystyle{splncs04}
\bibliography{main}

\newpage
\renewcommand{\thefigure}{S\arabic{figure}}
\renewcommand{\thetable}{S\arabic{table}}
\renewcommand{\theequation}{S\arabic{equation}}
\renewcommand{\thealgocf}{S\arabic{algocf}}
\renewcommand{\thesection}{S\arabic{section}}

\title{Supplementary Material to ``Imaging Interiors: An Implicit Solution to Electromagnetic Inverse Scattering Problems''} 

\titlerunning{An Implicit Solution to Electromagnetic Inverse Scattering Problems}

\author{Ziyuan Luo\inst{1}\orcidlink{0000-0003-1580-9809} \and
Boxin Shi\inst{2, 3}\orcidlink{0000-0001-6749-0364} \and
Haoliang Li\inst{4}\orcidlink{0000-0002-8723-8112} \and
Renjie Wan\inst{1}\orcidlink{0000-0002-0161-0367}}

\authorrunning{Z.~Luo et al.}

\institute{Department of Computer Science, Hong Kong Baptist University \and
National Key Laboratory for Multimedia Information Processing, School of Computer Science, Peking University \and
National Engineering Research Center of Visual Technology, School of Computer Science, Peking University \and
Department of Electrical Engineering, City University of Hong Kong \\
\email{ziyuanluo@life.hkbu.edu.hk}, \email{shiboxin@pku.edu.cn}, \email{haoliang.li@cityu.edu.hk}, \email{renjiewan@hkbu.edu.hk}}
\maketitle

\section{Overview}
This supplementary document provides more discussions, reproduction details, and additional results that accompany the main paper:
\begin{itemize}

    \item \cref{model} discusses the detailed physical model of the Electromagnetic Inverse Scattering Problems (EISP).

    \item \cref{settings} provides details of the system settings for each dataset.

    \item \cref{reproduction} presents reproduction details and pseudocode of our method.
    
    \item \cref{results} provides additional results, including ablation studies on different backbones and variation loss, and additional qualitative and quantitative results.
\end{itemize}

\section{Detailed physical model of EISP}
\label{model}
To clarify the physical model of the EISP, we copy some key equations in the main paper here. The data equation describes the wave-scatterer interaction, which can be formulated as
\begin{equation}
{\mathbf{E}}^{\mathrm{t}}={\mathbf{E}}^{\mathrm{i}}+{{\mathbf{G}}}_D \cdot {\mathbf{J}},\label{seq:d1}
\end{equation}
where ${\mathbf{E}}^{\mathrm{t}}$, ${\mathbf{E}}^{\mathrm{i}}$, and ${\mathbf{J}}$ are the discrete total electric fields, incident electric fields, and induced current, respectively. ${{\mathbf{G}}}_D$ is discrete free space Green’s function in Region of Interest (ROI) $D$. The relationship between the induced current $\mathbf{J}$ and total electric fields  ${\mathbf{E}}^{\mathrm{t}}$ can be described as
\begin{equation}
{\mathbf{J}} = {\text{Diag}(\boldsymbol{\xi})} \cdot {\mathbf{E}}^{\mathrm{t}},
\label{seq:d3}
\end{equation}
where $\text{Diag}(\boldsymbol{\xi})$ is the diagonal matrix of the contrast function. The contrast $\boldsymbol{\xi}$ is defined as
\begin{equation}
    \boldsymbol{\xi} = \pmb{\varepsilon_r}-1,
    \label{seq:xi_ep}
\end{equation}
where $\pmb{\varepsilon_r}$ is the relative permittivity. The data equation describes
the scattered ﬁeld as a reradiation of the induced current, which can be expressed as
\begin{equation}
{\mathbf{E}}^{\mathrm{s}}={{\mathbf{G}}}_S \cdot {\mathbf{J}},
\label{seq:d2}
\end{equation}
where ${\mathbf{E}}^{\mathrm{s}}$ is the discrete scattered ﬁeld, and  ${{\mathbf{G}}}_S$ is the discrete Green’s function to map the induced current ${\mathbf{J}}$ to scattered field ${\mathbf{E}}^{\mathrm{s}}$. 

\subsection{Forward estimation}
The aim of forward estimation is to deduce the scattered ﬁelds ${\mathbf{E}}^{\mathrm{s}}$ from given incident fields ${\mathbf{E}}^{\mathrm{i}}$. The forward estimation is linear because ${\mathbf{E}}^{\mathrm{s}}$ and ${\mathbf{E}}^{\mathrm{i}}$ have a linear relationship~\cite{chen2018computational}. Specifically, by replacing ${\mathbf{J}}$ in \cref{seq:d1} with \cref{seq:d3}, we can obtain
\begin{equation}
{\mathbf{E}}^{\mathrm{t}}={\mathbf{E}}^{\mathrm{i}}+{{\mathbf{G}}}_D \cdot  {\text{Diag}(\boldsymbol{\xi})} \cdot {\mathbf{E}}^{\mathrm{t}}. \label{seq:sub1}
\end{equation}
Reformulating \cref{seq:sub1} yields the expression for total fields ${\mathbf{E}}^{\mathrm{t}}$:
\begin{equation}
{\mathbf{E}}^{\mathrm{t}} = \left(\mathbf{I}-\mathbf{G}_D \text{Diag}(\boldsymbol{\xi})\right)^{-1} \cdot \mathbf{E}^{\text{i}}. \label{seq:sub2}
\end{equation}
By combining \cref{seq:d3}, we can obtain the expression of induced current ${\mathbf{J}}$ as
\begin{equation}
{\mathbf{J}} = \text{Diag}(\boldsymbol{\xi}) \cdot \left(\mathbf{I}-\mathbf{G}_D \text{Diag}(\boldsymbol{\xi})\right)^{-1} \cdot \mathbf{E}^{\text{i}}. \label{seq:sub3}
\end{equation}
Then, the expression for the scattered fields ${\mathbf{E}}^{\mathrm{s}}$ can be obtained from \cref{seq:d2} and \cref{seq:sub3} as
\begin{equation}
{\mathbf{E}}^{\mathrm{s}} = {{\mathbf{G}}}_S \cdot \text{Diag}(\boldsymbol{\xi}) \cdot \left(\mathbf{I}-\mathbf{G}_D \text{Diag}(\boldsymbol{\xi})\right)^{-1} \cdot \mathbf{E}^{\text{i}}. \label{seq:inv}
\end{equation}
Since Green’s functions ${\mathbf{G}}_D$ and ${\mathbf{G}}_S$ are fixed in our problem, and the contrast $\boldsymbol{\xi}$, or relative permittivity $\pmb{\varepsilon_r}$ is the physical property independent of the incident fields,~\cref{seq:inv} is a \underline{linear} equation in variables $\mathbf{E}^{\text{s}}$ and $\mathbf{E}^{\text{i}}$. Therefore, we can easily obtain the scattered fields $\mathbf{E}^{\text{s}}$ through \cref{seq:inv} if the relative permittivity $\pmb{\varepsilon_r}$ is known. We propose to make use of the convenience and benefits of the forward estimation to circumvent the difficulties of EISP. 

\subsection{Difficulties of EISP} In this section, we discuss three main challenges of EISP and explain why our approach can address these challenges. 

\paragraph{\textbf{Inverse.}} In an inverse problem, the incident fields ${\mathbf{E}}^{\mathrm{i}}$ are given, and the scattered ﬁelds $\mathbf{E}^{\text{s}}$ are measured by receivers, and then the task is to reconstruct relative permittivity $\pmb{\varepsilon_r}$ from the measured scattered fields. From~\cref{seq:xi_ep}, this task is equivalent to predicting the contrast $\boldsymbol{\xi}$. An intuitive approach is to infer the induced current ${\mathbf{J}}$ from the scattered field ${\mathbf{E}}^{\mathrm{s}}$ by inverse deduction from \cref{seq:d2}. However, the discrete Green’s function is a complex matrix of dimension $N_{\text{r}}\times M^2$, where $N_{\text{r}}$ is the total number of receivers and $M\times M$ is the size of the discretized subunits of ROI. In practice, we have $N_{\text{r}} \ll M^2$. Since such a less-than relation does not provide enough information to determine ${\mathbf{J}}$ from \cref{seq:d2}, it is difficult to obtain relative permittivity $\pmb{\varepsilon_r}$ in this inverse way.

\paragraph{\textbf{Nonlinearity.}} The nonlinearity poses significant challenges to the solution of the EISP. We explain nonlinearity from two perspectives. First, in \cref{seq:inv}, the nonlinearity is due to the fact that the scattered ﬁelds $\mathbf{E}^{\text{s}}$ are not doubled when the scatterer’s permittivity is doubled. This phenomenon is caused by the condition that total fields $E_t$ is a quantity related to the relative permittivity $\varepsilon_r$ according to \cref{seq:sub2}. Then, The nonlinearity is due to the multiple scattering effects that physically exist. In \cref{seq:d1}, the global-effect term ${{\mathbf{G}}}_D \cdot {\mathbf{J}}$ is caused by multiple scattering effects~\cite{chen2018computational}, a factor leading to the nonlinearity. Traditional methods, such as Born approximation~\cite{habashy1993beyond, 1132783}, involve a linearization of the original problem by neglecting the effect of multiple scattering. However, these methods can introduce significant errors and compromise the accuracy of the computation when the multiple scattering amplitude is large and unignorable.

\paragraph{\textbf{Discretization.}} Although the relative permittivity ${\varepsilon_r}$  exhibits continuous properties, numerical computations based on the aforementioned discrete equations can only obtain the discrete form of the relative permittivity with low resolution. Such a low resolution always makes it difficult to recognize the scatterer's details.

\paragraph{\textbf{Why can our approach address these challenges?}} We propose an implicit forward solution for EISP. First, we apply Implicit Neural Representations (INR) to represent relative permittivity ${\varepsilon_r}$ and induced current $J$ separately. Then we optimize these two representations through forward estimation by constructing two loss functions, namely data loss $\mathcal{L}_{\text{data}}$ and state loss $\mathcal{L}_{\text{state}}$. In this way, we do not need to worry about the difficulties caused by inverse estimation and nonlinearity. Besides, due to the inherent property of INR to approximate continuous functions, our method can provide results with flexible resolutions.

\section{Details of system settings}
\label{settings}
We conduct our experiments on Synthetic, real-world, and 3D datasets. There are some differences in system settings for each dataset, and we provide separate explanations for each.

\subsection{Settings for synthetic datasets}
\begin{figure}
  \centering
  \includegraphics[width=0.7\linewidth]{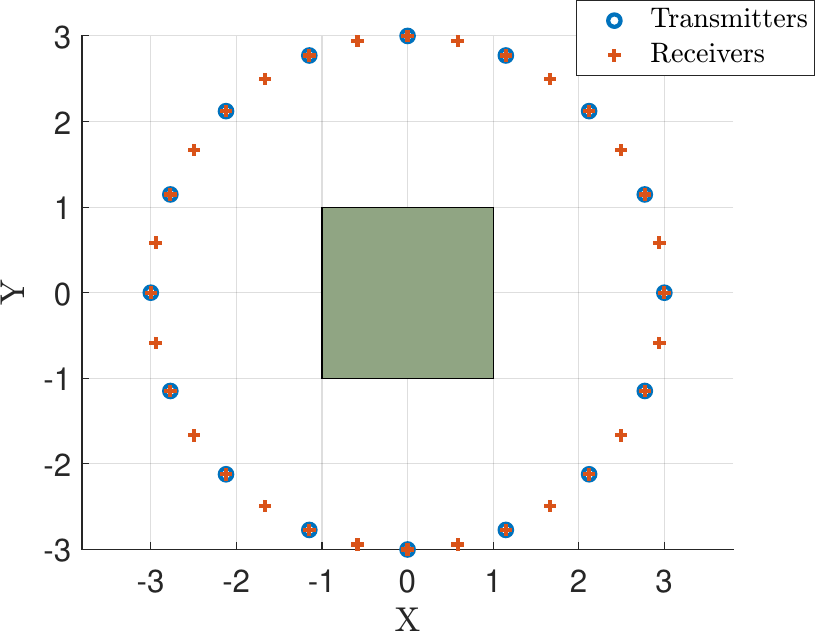}
  \caption{Positions of the transmitters and receivers on the measurement circle for synthetic datasets. The green area indicates the ROI.}
  \label{fig:2D_position}
  % \vspace{-10pt}
\end{figure}

Two synthetic datasets, the Circular-cylinder dataset and the MNIST dataset~\cite{wei2018deep, zhou2022deep}, are used for our experiments. The basic settings are the same for these two datasets. We set operating frequency $f = 400$ MHz, and the ROI is a square with the size of
$2\times2 \text{m}^2$. The placement scheme for transmitters and receivers is illustrated in~\cref{fig:2D_position}. There are $16$  transmitters and $32$ receivers equally placed on a circle of radius $3$ m centered at the center of ROI. 

\begin{figure*}
  \centering
  \includegraphics[width=0.8\linewidth]{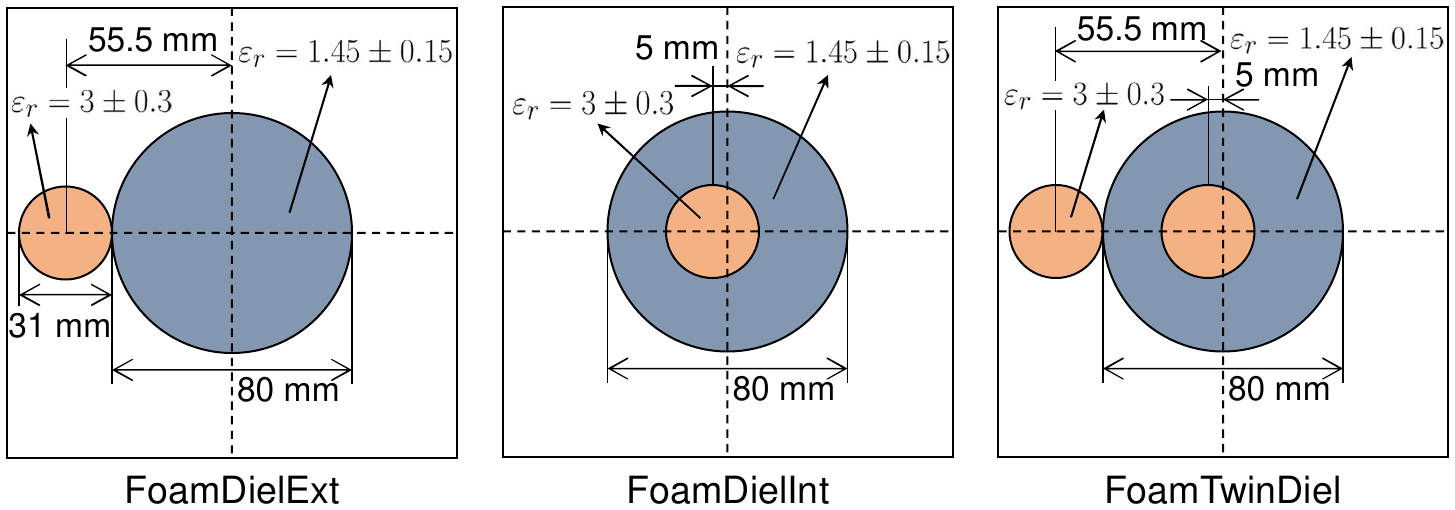}
  \caption{The ground truth of FoamDielExt, FoamDielInt, and FoamTwinDiel scenarios in Institut Fresnel’s database~\cite{geffrin2005free}.}
  \label{fig:real_world}
\end{figure*}

\begin{figure}
  \centering
  \includegraphics[width=0.5\linewidth]{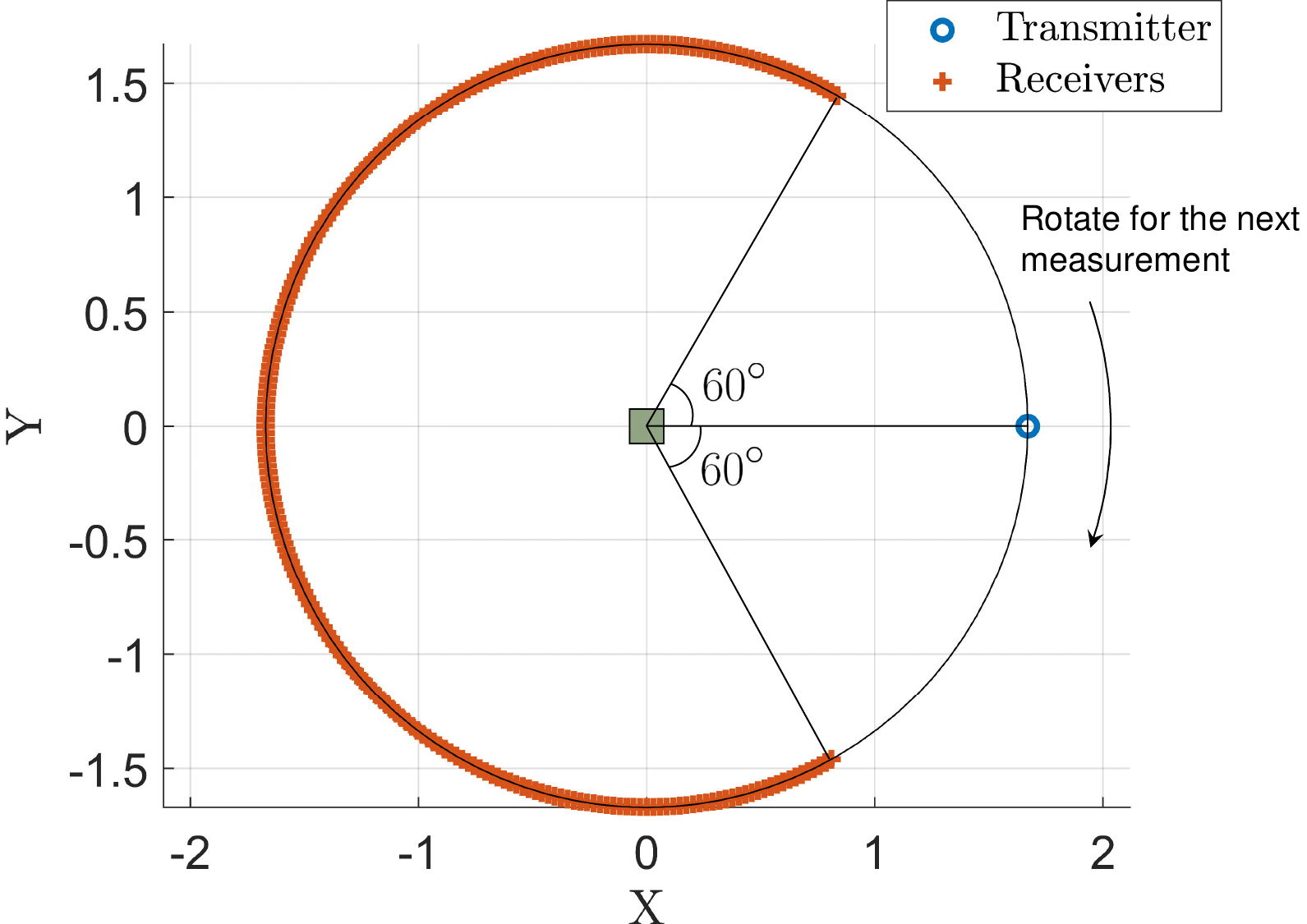}
  \caption{Positions of the transmitters and receivers on the measurement circle for Institut Fresnel’s database~\cite{geffrin2005free}. For FoamDielExt/FoamDielInt and FoamTwinDiel, the measurement system rotates by $45^{\circ}$ and $20^{\circ}$, respectively, for the next measurement. The green area indicates the ROI.}
  \label{fig:real_position}
\end{figure}

\subsection{Settings for real-world dataset}
Institut Fresnel’s database~\cite{geffrin2005free} is a famous real-world electromagnetic scattering dataset in the field of EISP. We use FoamDielExt, FoamDielInt, and FoamTwinDiel scenarios in Institut Fresnel’s database for testing. As presented in~\cref{fig:real_world}, all the cases consist of two kinds of cylinders. The large cylinder (SAITEC SBF 300) has a diameter of $80 \mathrm{~mm}$ with the relative permittivity $\varepsilon_r=1.45 \pm 0.15$. The small cylinder (berylon) has a diameter of $31 \mathrm{~mm}$ with the relative permittivity $\varepsilon_r=3 \pm 0.3$. The "$\pm$" indicates the range of uncertainty associated with the experimental value. The operating frequencies are taken from $2$ to $10$ GHz with a step of $1$ GHz. The ROI is a square with the size of
$0.15\times0.15 \text{m}^2$. All the transmitters and receivers are equally placed on a circle of radius $1.67$ m centered at the center of ROI. For all scenarios, $241$ receivers are used for each measurement, with a central angle step of $1^{\circ}$, without any position closer than $60^{\circ}$ from the transmitter~\cite{geffrin2005free}. The placement schemes for FoamDielExt, FoamDielInt, and FoamTwinDiel are shown in~\cref{fig:real_position}. After each measurement, the measurement system rotates by a certain angle for the next measurement. For FoamDielExt and FoamDielInt, this angle is $45^{\circ}$, while for FoamTwinDiel, it is $20^{\circ}$. This means there are $8$ transmitters for FoamDielExt and FoamDielInt, while there are $18$ transmitters for FoamTwinDiel.

\subsection{Settings for 3D dataset}
\begin{figure}
  \centering
  \includegraphics[width=0.45\linewidth]{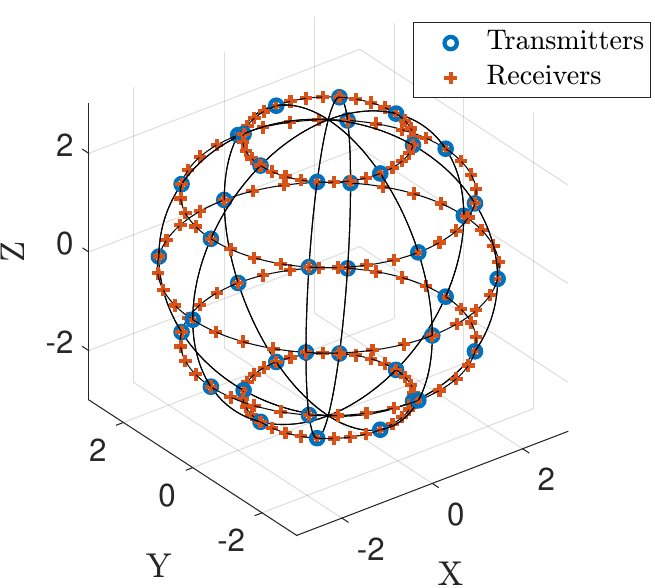}
  \caption{Positions of the transmitters and receivers on the measurement sphere for 3D dataset.}
  \label{fig:3D_position}
  % \vspace{-10pt}
\end{figure}

We also test our method on the 3D MNIST dataset~\cite{3Dmnist}. We set the permittivities of the objects to be $2$. We set operating frequency $f = 400$ MHz, and the ROI is a cube with the size of $2 \times 2\times 2 \text{m}^3$. As shown in~\cref{fig:3D_position}, there are $40$ transmitters and $160$ receivers. The transmitters and receivers are all located at the sphere of radius $3$ m around the target centered at the center of ROI. For the positions of transmitters, the azimuthal angle ranged from $0^{\circ}$ to $315^{\circ}$ with a $45^{\circ}$ step, and the polar angle ranged from $30^{\circ}$ to $150^{\circ}$ with a $30^{\circ}$ step. For the positions of receivers, the azimuthal angle ranged from $0^{\circ}$ to $348.75^{\circ}$ with an $11.25^{\circ}$ step, and the polar angle ranged from $30^{\circ}$ to $150^{\circ}$ with a $30^{\circ}$ step.

\section{Reproduction details and Pseudocode}
\label{reproduction}

\subsection{Reproduction details}
In this section, we present reproduction details and pseudocode of our method. Our code will be released upon the acceptance of this paper.
\paragraph{\textbf{Additional network details.}} Two eight-layer MLPs with $256$ channels and ReLU activations are used to individually predict the relative permittivity $\varepsilon_r$ and induced current $J$. The difference between these two networks lies in the last layer. The output dimension of the last layer is $1$ for relative permittivity and $2$ for induced current, representing the real and imaginary parts, respectively.

\paragraph{\textbf{Computational details.}} In Section 4.2, we have developed formulas to predict the relative permittivity and induced current for each transmitter. Specifically, for $p$-th transmitter, we can obtain the predicted scattered fields as
\begin{equation}
\hat{\mathbf{E}}^{\mathrm{s}}_p={{\mathbf{G}}}_S \cdot {\mathbf{J}_p},\  p=1,2,\cdots, N_{\text{t}},
\label{seq:E}
\end{equation}
where ${{\mathbf{G}}}_S$ is a $N_{\text{r}}\times M^2$ complex matrix, denoting the discrete Green's function, and $\mathbf{J}_p$ and $\hat{\mathbf{E}}^{\mathrm{s}}_p$ are complex vectors of dimensions $M^2$ and $N_{\text{r}}$, denoting the discrete induced current and predicted scattered fields inspired by $p$-th transmitter, respectively. $N_{\text{t}}$ and $N_{\text{r}}$ are the total numbers of transmitters and receivers, respectively, and $M\times M$ is the size of the discretized subunits of ROI. $\mathbf{J}_p$ is the discrete induced current directly sampled from $H_\phi$. To calculate the mismatch of the state equation, we can obtain the predicted induced current $\hat{\mathbf{J}}_p$ as
\begin{equation}
\hat{\mathbf{J}}_p = \text{Diag}(\boldsymbol{\xi})\cdot {\mathbf{E}}_p^{\text{i}}+\text{Diag}(\boldsymbol{\xi}) \cdot {{\mathbf{G}}}_D \cdot {\mathbf{J}}_p,\  p=1,2,\cdots, N_{\text{t}},
\label{seq:J}
\end{equation}
where $\boldsymbol{\xi}$ is a vector of dimension $M^2$, reshaped from spatially queried network $F_\theta$, representing the contrast value. $\hat{\mathbf{J}}_p$ is a complex vector of dimension $M^2$, indicating the induced current computed via the mathematical correlation. $\text{Diag}(\boldsymbol{\xi})$ is the diagonal matrix of the contrast function with dimension $M^2 \times M^2$. ${{\mathbf{G}}}_D$ is also a discrete Green's function with dimension $M^2 \times M^2$.

Although we provide the computation formulas for each transmitter when calculating \cref{seq:E} and \cref{seq:J} for all $N_{\text{t}}$ transmitters, we use a more efficient approach. To be precise, equations in \cref{seq:E} and \cref{seq:J} can be rewritten as 
\begin{equation}
\hat{\mathbf{E}}^{\mathrm{s}}_{\text{all}} ={{\mathbf{G}}}_S \cdot {\mathbf{J}_{\text{all}}},
\label{seq:E_2}
\end{equation}
and
\begin{equation}
\hat{\mathbf{J}}_{\text{all}} = \text{Diag}(\boldsymbol{\xi})\cdot {\mathbf{E}}_{\text{all}}^{\text{i}} + \text{Diag}(\boldsymbol{\xi}) \cdot {{\mathbf{G}}}_D \cdot {\mathbf{J}}_{\text{all}},
\label{seq:J_2}
\end{equation}
where
\begin{equation}
\hat{\mathbf{E}}^{\mathrm{s}}_{\text{all}} = [\hat{\mathbf{E}}^{\mathrm{s}}_1, \hat{\mathbf{E}}^{\mathrm{s}}_2, 
\cdots, \hat{\mathbf{E}}^{\mathrm{s}}_{N_{\text{t}}}],
\end{equation}
\begin{equation}
    {\mathbf{J}}_{\text{all}} = [{\mathbf{J}_1}, {\mathbf{J}_2}, \cdots, {\mathbf{J}_{N_{\text{t}}}}],
\end{equation}
\begin{equation}
    \hat{\mathbf{J}}_{\text{all}} = [\hat{\mathbf{J}}_1, \hat{\mathbf{J}}_2, \cdots, \hat{\mathbf{J}}_{N_{\text{t}}}],
\end{equation}
\begin{equation}
    {\mathbf{E}}_{\text{all}}^{\text{i}} = [{\mathbf{E}}_1^{\text{i}}, {\mathbf{E}}_2^{\text{i}}, \cdots, {\mathbf{E}}_{N_{\text{t}}}^{\text{i}}].
\end{equation}
In \cref{seq:E_2} and \cref{seq:J_2}, $\hat{\mathbf{E}}^{\mathrm{s}}_{\text{all}}$ is a matrix of dimension $N_{\text{r}}\times N_{\text{t}}$, containing the scattered fields referring to all transmitters.  ${\mathbf{J}}_{\text{all}}$, $\hat{\mathbf{J}}_{\text{all}}$, and ${\mathbf{E}}_{\text{all}}^{\text{i}}$ are all $M^2\times N_{\text{t}}$ matrices. Therefore, during implementation, the loss functions can be equivalently rewritten as
\begin{equation}
\mathcal{L}_{\text{data}} = \frac{\|\hat{\mathbf{E}}^{\text{s}}_{\text{all}}-{\mathbf{E}}^{\text{s}}_{\text{all}}\|_2^2}{\|\mathbf{E}^{\text{s}}_{\text{all}}\|_2^2},\label{seq:loss_data}
\end{equation}
and
\begin{equation}
\mathcal{L}_{\text{state}} = \frac{\|\hat{\mathbf{J}}_{\text{all}}-{\mathbf{J}}_{\text{all}}\|_2^2}{\|\ \text{Diag}(\boldsymbol{\xi})\cdot {\mathbf{E}}_{\text{all}}^{\text{i}}\|_2^2 + \Delta},\label{seq:loss_state}
\end{equation}
where ${\mathbf{E}}^{\text{s}}_{\text{all}}$ is the ground truth measured by receivers in a matrix form, $\Delta$ denotes a small number to improve stability by preventing the denominator from being zero, and $\|\cdot\|_2$ denotes the matrix $2$-norm.

\paragraph{\textbf{Calculation of Green's functions.}}
The two-dimensional scalar Green’s function~\cite{chen2018computational} can be expressed as
\begin{equation}
g\left(\mathbf{x}, \mathbf{x}^{\prime}\right)=\frac{i}{4} H_0^{(1)}\left(k_0\left|\mathbf{x}-\mathbf{x}^{\prime}\right|\right),
\end{equation}
where $H_0^{(1)}(\cdot)$ is the zeroth order Hankel function of the first kind, $k_0=2 \pi / \lambda_0$ is the wavenumber in free space, and $\lambda_0$ is the wavelength in free space. $\mathbf{x}$ and $\mathbf{x}^{\prime}$ are the coordinates of two corresponding positions.

We use the method of moment (MOM)~\cite{peterson1998computational} with the pulse basis function and the delta testing function to discretize the domain $D$ into $M \times M$ subunit, and the
centers of subunits are located at $\mathbf{x}_n$ with $n = 1, 2, ..., M^2$. Then, we can discretize this continuous Green's function into matrix ${{\mathbf{G}}}_D$ and ${{\mathbf{G}}}_S$ respectively. The element in the $n-$th row and $n^{\prime}-$th column of the $M \times M$ matrix ${{\mathbf{G}}}_D$ can be obtained as
\begin{equation}
{{\mathbf{G}}}_D(n, n^{\prime}) = k_0^2 A_{n^{\prime}} g\left(\mathbf{x}_n, \mathbf{x}_{n^{\prime}}\right),\ n=1,2, \ldots, M^2, n^{\prime}=1,2, \ldots, M^2,
\label{eq:GD}
\end{equation}
where $A_{n^{\prime}}$ is the area of the $n^{\prime}-$th subunits. Similarly, the element in the $q-$th row and $n^{\prime}-$th column of the $N_{\text{r}} \times M$ matrix ${{\mathbf{G}}}_S$ can be obtained as
\begin{equation}
{{\mathbf{G}}}_S(q, n^{\prime}) = k_0^2 A_{n^{\prime}} g\left(\mathbf{x}_q, \mathbf{x}_{n^{\prime}}\right),\ q=1,2, \ldots, N_\text{t}, n^{\prime}=1,2, \ldots, M^2.
\label{eq:GS}
\end{equation}
The discretized forms of Green's function can then be used in the calculations in~\cref{seq:E} to~\cref{seq:J_2}.

\paragraph{\textbf{Preprocessing for real-world dataset.}}
To handle real-world and synthetic data in a unified manner, we calibrate the real-world data before using it. Following previous works~\cite{wei2018deep, jin2017deep, liu2022physics}, the calibration of real-world scattering field data can be conducted by multiplying those data with a complex coefficient. The complex coefficient is derived by dividing the measured incident field by the simulated incident field at the receiver located opposite the source~\cite{geffrin2005free}.

\paragraph{\textbf{Implementation details of baselines.}}
For Physics-Net~\cite{liu2022physics}, we follow the formulation in~\cite{liu2022physics} to get the regularization parameter $\beta$. We use the backbone architecture depicted in the same paper. For network optimization, we use the SDG optimizer with momentum $0.99$, a learning rate $5 \times 10^{-6}$ that decays following the step scheduler with step size $20$ and decay factor $0.5$. All the hyperparameters are recommended in the paper.

For PGAN~\cite{9468919}, the structure of the generator and discriminator follows the architecture in~\cite{9468919}, respectively. We also use the hyperparameters suggested in the paper. The number of hidden layers used in perceptual adversarial loss is $M_d=1$, weight parameters $\beta=0.01$ and $\gamma=4.0$ for the loss function of the generator, and $m=0.2$ for the loss function of the discriminator. For network optimization, we employ the Adam optimizer with default values $\beta_1=0.9, \beta_2=0.999, \epsilon=10^{-8}$, and a learning rate $2 \times 10^{-4}$ that decays following the linear scheduler after the first $20$ epochs during optimization. All the hyperparameters are the ones suggested by the paper.

We directly use the codes of BPS and CS-Net to ensure the fairness of the evaluation.

\subsection{Pseudocode}
We provide a pseudocode to offer a detailed and step-by-step understanding of our proposed approach, as shown in~\cref{alg:algorithm}.

\begin{algorithm}[!htb]
\caption{Our INR-based method for EISP}\label{alg:algorithm}
\KwData{Incident field $\mathbf{E}^{\text{t}}$, scattered field $\mathbf{E}^{\text{s}}$, ROI $D$, transmitters' positions $\mathbf{x}^{\text{t}}$, and receivers' positions $\mathbf{x}^{\text{r}}$}
\KwResult{Optimized INR $F_\theta$ for the object's relative permittivity}
Initial INR $F_\theta$ for relative permittivity, and INR $H_\phi$ for induced current;

Generate $\mathbf{G}_D$ and $\mathbf{G}_S$ from $\mathbf{x}^{\text{t}}$, $\mathbf{x}^{\text{r}}$ and $D$ according to definition of discretized forms of Green's function from~\cref{eq:GD} and~\cref{eq:GS}, respectively;

\textbf{For} $\text{step}=1$ \textbf{to} $\text{max\_iter}$:\\
\quad Infer contrast $\boldsymbol{\xi}$ from $F_\theta$ using $D$, and infer induced current $\mathbf{J}$ from $H_\phi$ using $D$ and $\mathbf{x}^{\text{t}}$ with random spatial sampling\;
\quad Calculate $\hat{\mathbf{E}}^{\text{s}}$ and $\hat{\mathbf{J}}$ from~\cref{seq:E_2} and~\cref{seq:J}, respectively\;
\quad Calculate $\mathcal{L}_{\text{data}}$, $\mathcal{L}_{\text{state}}$ and $\mathcal{L}_{\text{TV}}$ from~\cref{seq:loss_data},~\cref{seq:loss_state} and total variation function, respectively\;
\quad Obtain the loss $\mathcal{L} = \lambda_{\text{data}}\mathcal{L}_{\text{data}} + \lambda_{\text{state}}\mathcal{L}_{\text{state}} + \lambda_{\text{TV}}\mathcal{L}_{\text{TV}}$\;
\quad Update $\theta$ and $\phi$ by minimizing $\mathcal{L}$ using the Adam optimizer.
\end{algorithm}

\section{Additional results}
\label{results}

\begin{figure*}[ht]
  \centering
  \includegraphics[width=.85\linewidth]{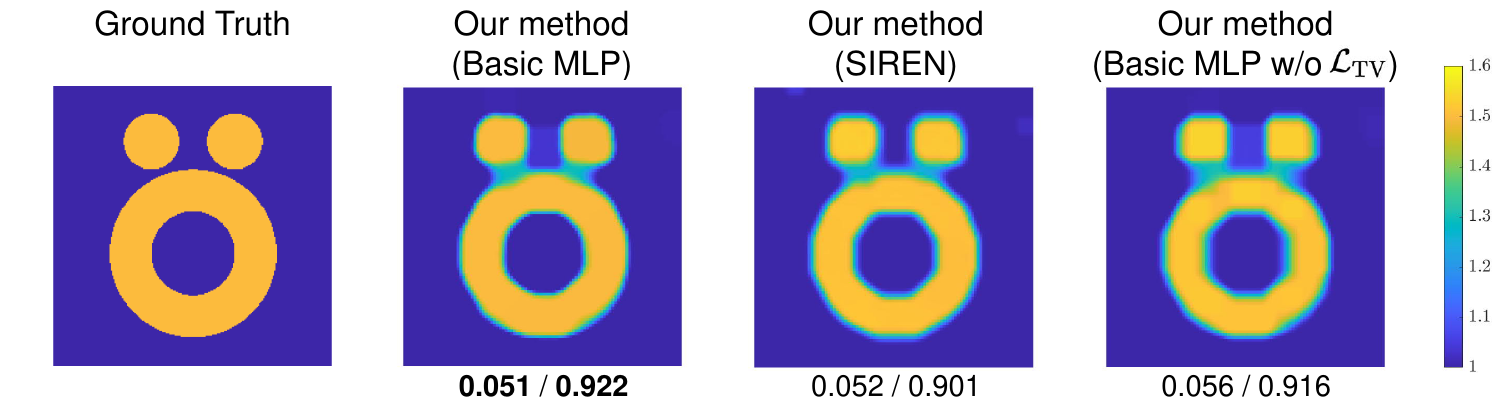}
  \caption{Ablation study results including comparison with different backbones and comparison with and without variation loss $\mathcal{L}_{\text{TV}}$. We show the results of a standard test case~\cite{belkebir1996using} in EISP. The pixel values in the images indicate the values of the relative permittivity. RRMSE/SSIM values are shown below the figure.}
  \label{fig:tv_ablation}
\end{figure*}

\paragraph{\textbf{Ablation study on different backbones.}}
We study two different backbones for INR, namely basic MLP with ReLU activations and SIREN~\cite{sitzmann2020implicit}. These two structures are both based on fully connected networks to represent continuous mappings, so choosing either network does not affect our proof of the applicability of INR. The results for different backbones are shown in~\cref{fig:tv_ablation}. From the results, both basic MLP and SIREN can accurately reconstruct the internal structures of objects. The reconstruction quality using basic MLP is slightly better than that of SIREN.

Some previous studies point out that SIREN~\cite{sitzmann2020implicit} has certain drawbacks in terms of its implementation~\cite{conde2024nilut}.
First, it cannot utilize the speed-up techniques of INRs, such as the one described in Instant-NGP~\cite{muller2022instant}. Second, their custom activations are still not compatible with accelerator hardware in certain devices~\cite{conde2024nilut}. Therefore, we choose the basic MLP as the backbone of INR in our main paper.

\paragraph{\textbf{Ablation study on variation loss.}} We further test the impact of total variation loss $\mathcal{L}_{\text{TV}}$ for relative permittivity $\boldsymbol{\xi}$ on the results. We show the results with and without $\mathcal{L}_{\text{TV}}$ in \cref{fig:tv_ablation}. The results indicate that $\mathcal{L}_{\text{TV}}$ improves our model's performance.

\begin{figure*}
  \centering
  \includegraphics[height=\textheight-4.0\baselineskip]{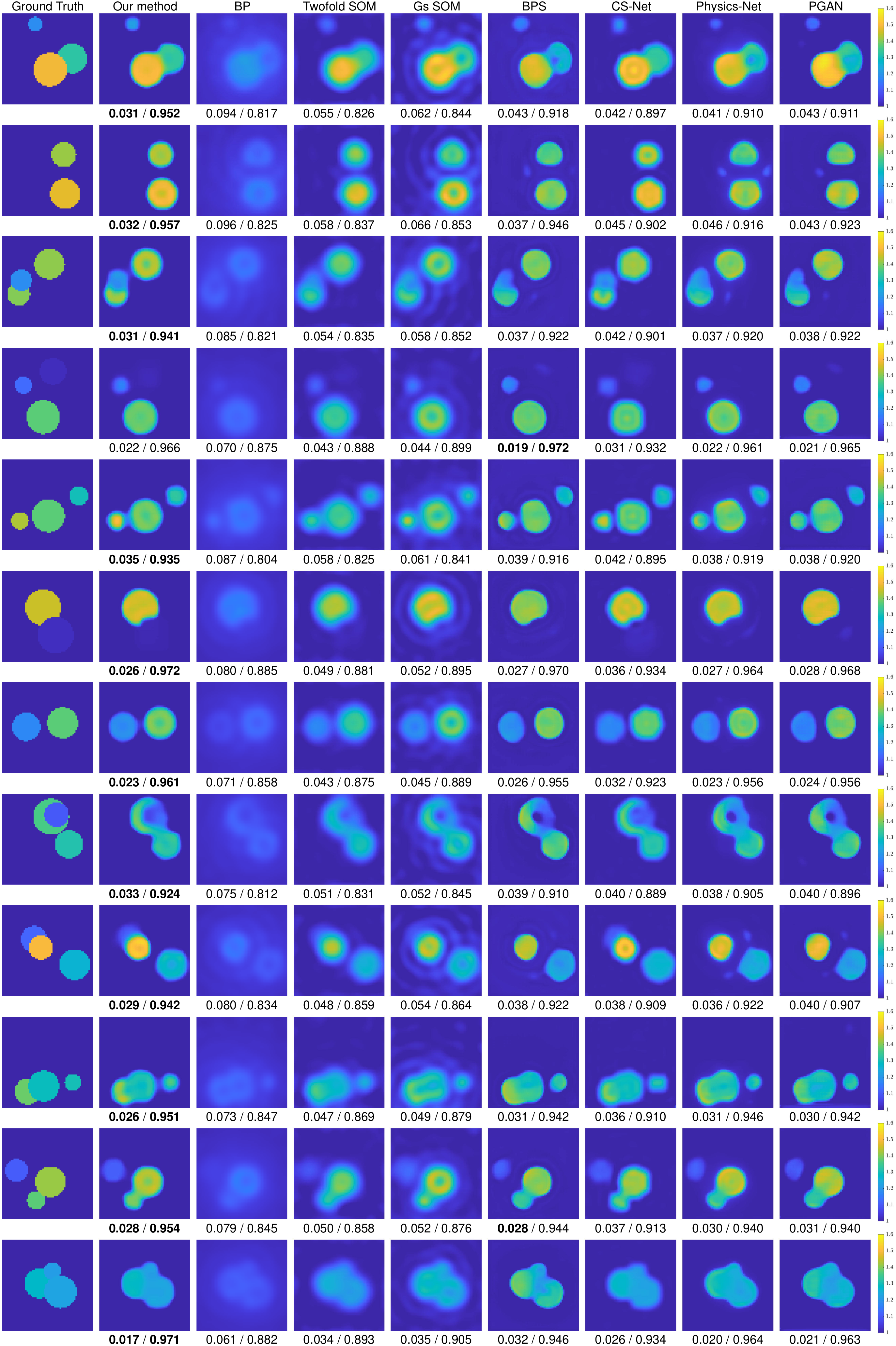}
  \caption{Samples obtained from synthetic Cicrular-cylinder dataset under $5\%$ noise level. From left to right: ground truth, results obtained by our method, {BP}~\cite{belkebir2005superresolution}, {Twofold SOM}~\cite{zhong2009twofold},  {Gs SOM}~\cite{chen2009subspace}, {BPS}~\cite{wei2018deep}, {CS-Net}~\cite{sanghvi2019embedding}, 
 {Physics-Net}~\cite{liu2022physics}, and {PGAN}~\cite{9468919}. RRMSE/SSIM values are shown below each figure.}
  \label{fig:Circle_0_05}
  % \vspace{-10pt}
\end{figure*}

\begin{figure*}
  \centering
  \includegraphics[height=\textheight-4.0\baselineskip]{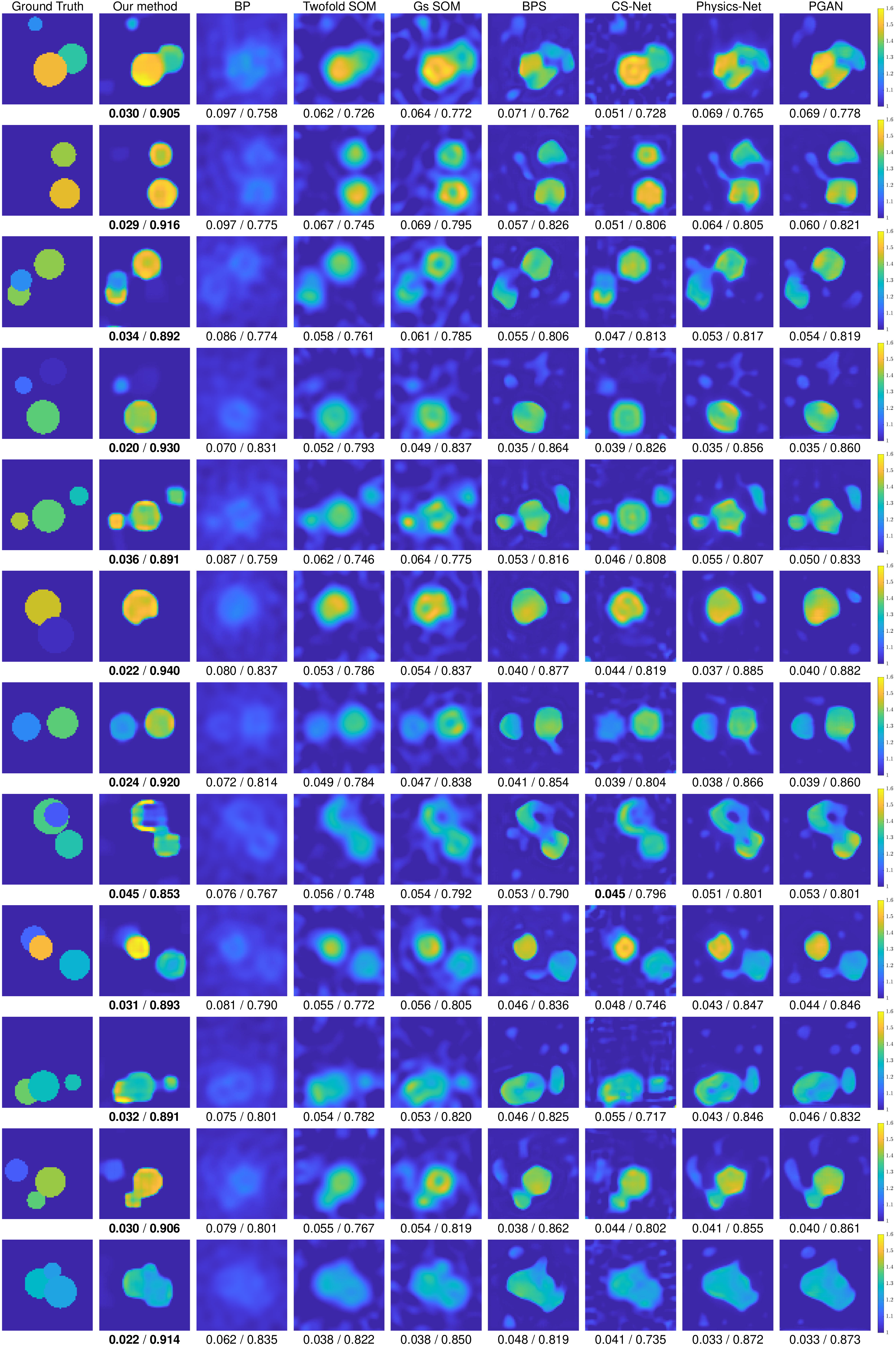}
  \caption{Samples obtained from synthetic Cicrular-cylinder dataset under $30\%$ noise level. From left to right: ground truth, results obtained by our method, {BP}~\cite{belkebir2005superresolution}, {Twofold SOM}~\cite{zhong2009twofold},  {Gs SOM}~\cite{chen2009subspace}, {BPS}~\cite{wei2018deep}, {CS-Net}~\cite{sanghvi2019embedding}, 
 {Physics-Net}~\cite{liu2022physics}, and {PGAN}~\cite{9468919}. RRMSE/SSIM values are shown below each figure.}
  \label{fig:Circle_0_3}
  % \vspace{-10pt}
\end{figure*}

\begin{figure*}
  \centering
  \includegraphics[height=\textheight-4.0\baselineskip]{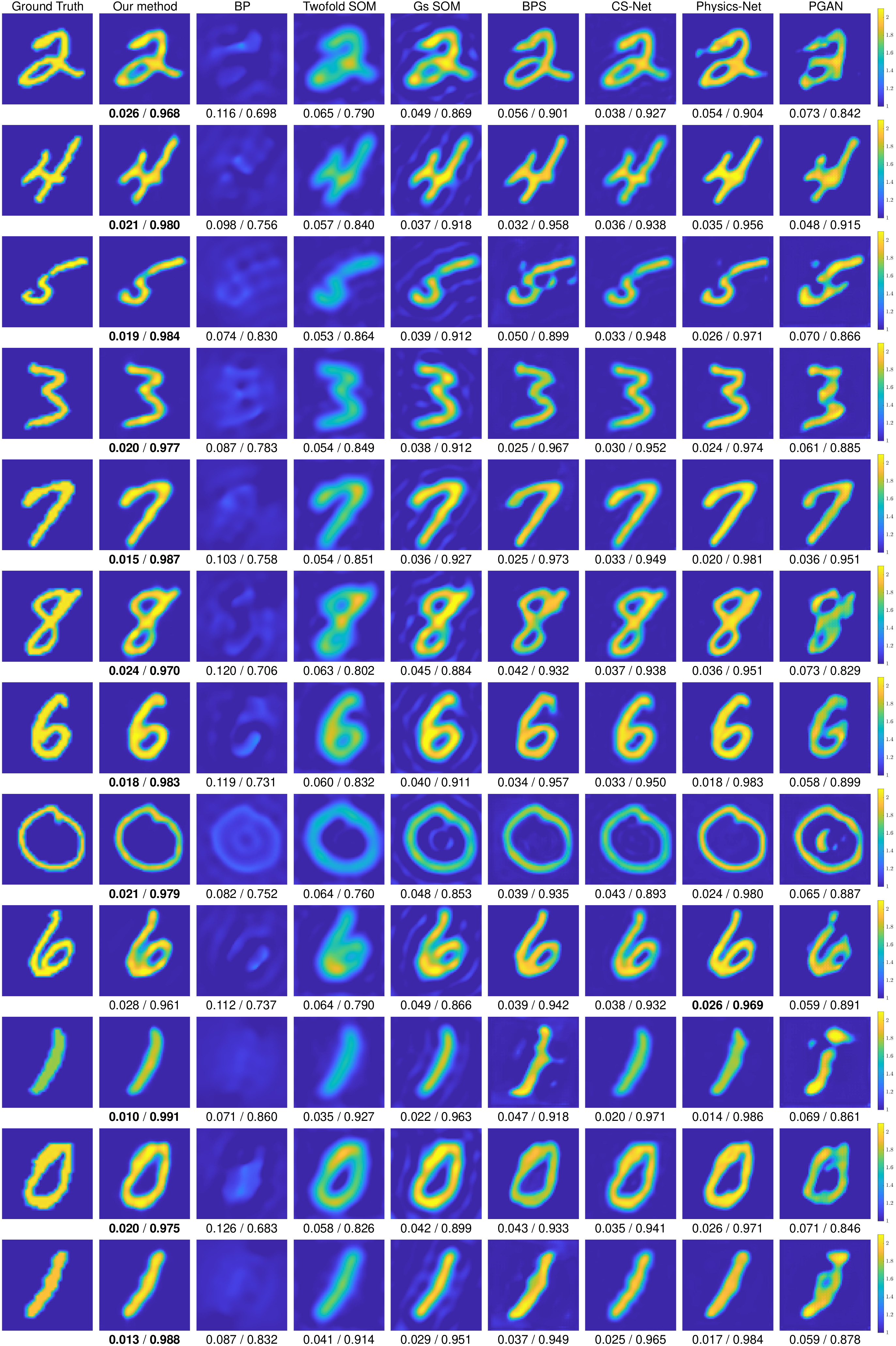}
  \caption{Samples obtained from synthetic MNIST dataset under $5\%$ noise level. From left to right: ground truth, results obtained by our method, {BP}~\cite{belkebir2005superresolution}, {Twofold SOM}~\cite{zhong2009twofold},  {Gs SOM}~\cite{chen2009subspace}, {BPS}~\cite{wei2018deep}, {CS-Net}~\cite{sanghvi2019embedding}, 
 {Physics-Net}~\cite{liu2022physics}, and {PGAN}~\cite{9468919}. RRMSE/SSIM values are shown below each figure.}
  \label{fig:Mnist_0_05}
  % \vspace{-10pt}
\end{figure*}

\begin{figure*}
  \centering
  \includegraphics[height=\textheight-4.0\baselineskip]{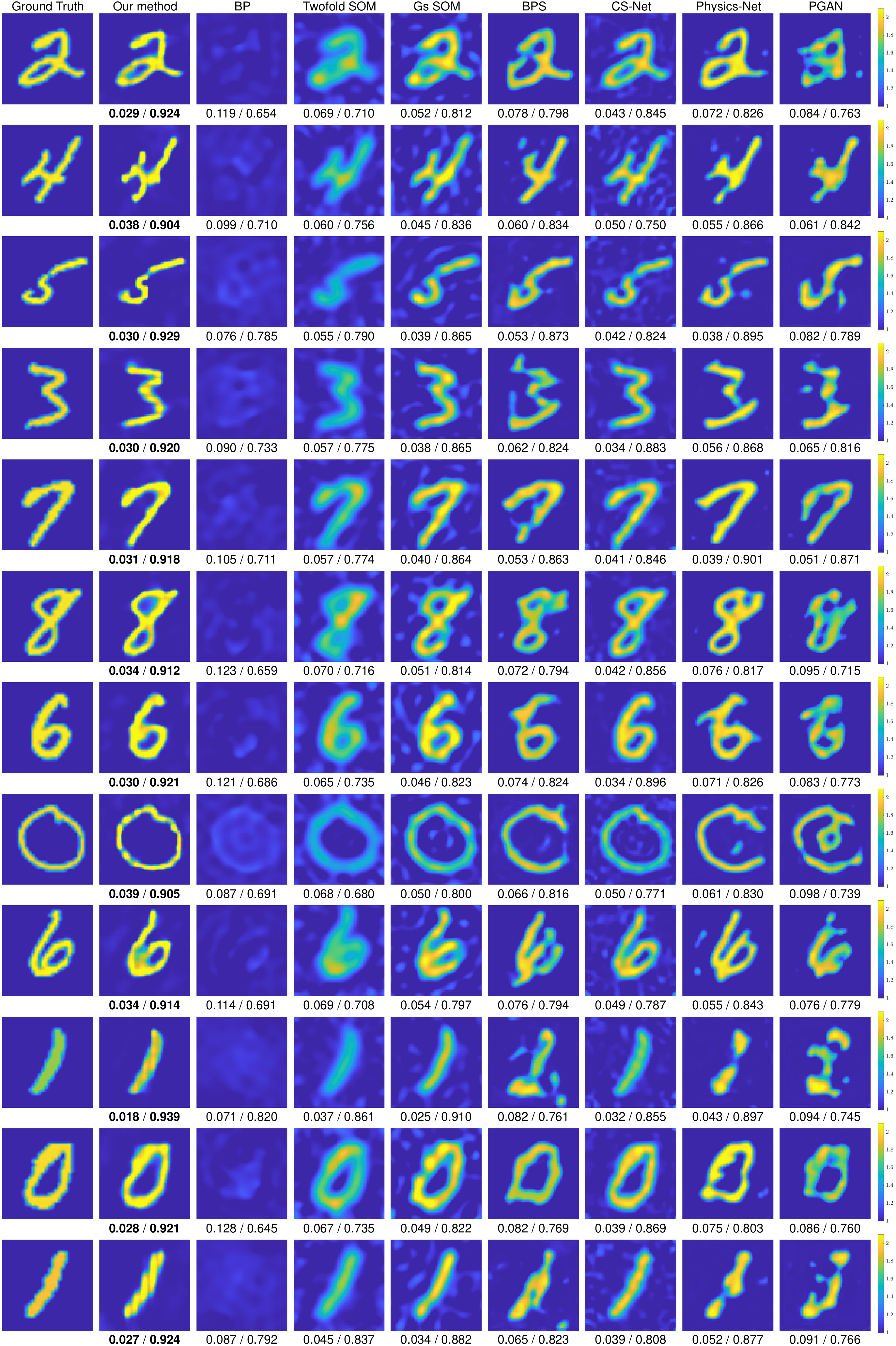}
  \caption{Samples obtained from synthetic MNIST dataset under $30\%$ noise level. From left to right: ground truth, results obtained by our method, {BP}~\cite{belkebir2005superresolution}, {Twofold SOM}~\cite{zhong2009twofold},  {Gs SOM}~\cite{chen2009subspace}, {BPS}~\cite{wei2018deep}, {CS-Net}~\cite{sanghvi2019embedding}, 
 {Physics-Net}~\cite{liu2022physics}, and {PGAN}~\cite{9468919}. RRMSE/SSIM values are shown below each figure.}
  \label{fig:Mnist_0_3}
  % \vspace{-10pt}
\end{figure*}

\paragraph{\textbf{Additional qualitative and quantitative results.}} We present more qualitative and quantitative results on the Circular-cylinder dataset~\cite{wei2018deep} and MNIST dataset~\cite{deng2012mnist} under different noise levels, as shown in~\cref{fig:Circle_0_05} to \cref{fig:Mnist_0_3}. Our method reaches the highest visual quality compared with the other baseline methods.

\end{document}